\documentclass[10pt,journal,compsoc]{IEEEtran}

\ifCLASSOPTIONcompsoc
  \usepackage[nocompress]{cite}

\else
  \usepackage{cite}

\fi

\ifCLASSINFOpdf
\else
\fi

\usepackage{caption}

\usepackage{url}
\usepackage{lmodern}
\usepackage{times}
\usepackage{epsfig}
\usepackage{graphicx}
\usepackage{amsmath}
\usepackage{amssymb}
\usepackage{textcomp}

\usepackage{multirow}
\usepackage{colortbl}
\usepackage{color}
\usepackage{threeparttable}
\usepackage{booktabs}
\usepackage{adjustbox}
\usepackage{subfig}
\usepackage[misc]{ifsym}
\usepackage{xcolor,colortbl}
\usepackage{makecell}
\usepackage{graphicx}
\usepackage{bm}
\usepackage{tikz}
\usetikzlibrary{arrows.meta, positioning, shapes.geometric, calc}
\usepackage{xcolor}
\usepackage{amsmath}
\usepackage{booktabs}
\usepackage{colortbl}
\newcommand{\revise}[1]{{\color{black}#1}}

\newcommand*{\belowrulesepcolor}[1]{%
  \noalign{%
    \kern-\belowrulesep 
    \begingroup 
      \color{#1}%
      \hrule height\belowrulesep 
    \endgroup 
    \vspace{-0.03mm}
  }%
} 
\newcommand*{\aboverulesepcolor}[1]{%
  \noalign{%
  \vspace{-0.03mm}
    \begingroup 
      \color{#1}%
      \hrule height\aboverulesep 
    \endgroup 
    \kern-\aboverulesep 
  }%
}

\usepackage{bbding}
\usepackage{pifont}
\usepackage{stfloats}
\usepackage{wasysym}
\newcommand{\cmarkg}{\textcolor{lightgray}{\ding{51}}\xspace}%
\newcommand{\xmark}{\ding{55}\xspace}%
\usepackage{amsmath,amssymb}

\usepackage{pifont}

\usepackage{xspace}
\makeatletter
\DeclareRobustCommand\onedot{\futurelet\@let@token\@onedot}
\def\@onedot{\ifx\@let@token.\else.\null\fi\xspace}
\def\eg{\emph{e.g}\onedot}

\def\etal{\emph{et al}\onedot}

\usepackage[linesnumbered,lined,boxed,commentsnumbered,ruled]{algorithm2e}
\usepackage[breaklinks=true,colorlinks,bookmarks=false]{hyperref}
\def\etal{\textit{et al.}}
\definecolor{mygray}{gray}{.85}
\definecolor{mygray1}{gray}{.7}
\definecolor{mygray2}{gray}{.93}

\usepackage[T1]{fontenc}
\newcommand{\best}[1]{\textcolor{black}{\textbf{#1}}}

\newcommand{\thickhline}{
    \noalign {\ifnum 0=`}\fi \hrule height 1pt
    \futurelet \reserved@a \@xhline
}
\newcommand{\middlefont}{\fontsize{6pt}{7pt}\selectfont}
\newcommand{\pub}[1]{\color{gray}{\scriptsize{[{#1}]}}}
\newcommand{\mypub}[1]{\color{gray}{\footnotesize{[{#1}]}}}

\begin{document}

\title{A Survey on 3D Gaussian Splatting Applications: Segmentation, Editing, and Generation}

\author{Shuting He, Peilin Ji, Yitong Yang, Changshuo Wang, Jiayi Ji, Yinglin Wang, Henghui Ding
\IEEEcompsocitemizethanks{
\IEEEcompsocthanksitem S. He, P. Ji, Y. Yang, and Y. Wang are with Shanghai University of Finance and Economics, China. shuting.he@sufe.edu.cn.
\IEEEcompsocthanksitem C. Wang is with University College London, United Kingdom. 
\IEEEcompsocthanksitem J. Ji is with Xiamen University, China.
\IEEEcompsocthanksitem H. Ding is with Fudan University, China. henghui.ding@gmail.com.
}
}
\markboth{IEEE TRANSACTIONS ON PATTERN ANALYSIS AND MACHINE INTELLIGENCE}
{Shell \MakeLowercase{\textit{et al.}}: Bare Advanced Demo of IEEEtran.cls for IEEE Computer Society Journals}

\IEEEtitleabstractindextext{

\begin{abstract}
In the context of novel view synthesis, 3D Gaussian Splatting (3DGS) has recently emerged as an efficient and competitive counterpart to Neural Radiance Field (NeRF), enabling high-fidelity photorealistic rendering in real time.Beyond novel view synthesis, the explicit and compact nature of 3DGS enables a wide range of downstream applications that require geometric and semantic understanding. This survey provides a comprehensive overview of recent progress in 3DGS applications. 
It first reviews the reconstruction preliminaries of 3DGS, followed by the problem formulation, 2D foundation models, and related NeRF-based research areas that inform downstream 3DGS applications.
We then categorize 3DGS applications into three foundational tasks: segmentation, editing, and generation, alongside additional functional applications built upon or tightly coupled with these foundational capabilities.
For each, we summarize representative methods, supervision strategies, and learning paradigms, highlighting shared design principles and emerging trends. Commonly used datasets and evaluation protocols are also summarized, along with a comparative analysis of recent methods across public benchmarks. To support ongoing research and development, a continually updated repository of papers, code, and resources is maintained at
\url{https://github.com/heshuting555/Awesome-3DGS-Applications}.
\end{abstract}

}

\maketitle                                                                                    

\IEEEdisplaynontitleabstractindextext

\IEEEpeerreviewmaketitle

\section{Introduction}
\label{sec:introduction}

\IEEEPARstart{3}{D} Gaussian Splatting (3DGS)\cite{3DGS} has recently emerged as a powerful paradigm for real-time neural rendering, achieving high-fidelity photorealistic synthesis with superior efficiency. Neural Radiance Field (NeRF)~\cite{NeRF} represents a scene as an implicit radiance field parameterized by a neural network that maps 3D coordinates and viewing directions to density and view-dependent radiance, which is rendered via volume rendering. In contrast, 3DGS represents a scene as an explicit set of anisotropic 3D Gaussians and renders them via differentiable rasterization. This explicit yet learnable formulation enables efficient optimization and fast inference, while preserving fine-grained geometric and appearance details. As a next-generation explicit 3D representation, 3DGS has demonstrated remarkable potential across a broad spectrum of applications, including virtual and augmented reality, robotics, autonomous navigation, and urban mapping.

While early research on 3DGS~\cite{3DGS, Scaffold-gs, chen2024mvsplat} primarily focused on novel view synthesis, recent works have extended its scope to a growing number of downstream tasks, such as simultaneous localization and mapping (SLAM)\cite{yan2024gs}, human avatar~\cite{GaussianAvatars}, segmentation~\cite{Ref-LERF}, editing~\cite{GaussianEditor-HGS}, generation~\cite{Dreamgaussian}, and more. These applications require richer representations that go beyond geometry to incorporate semantics, spatial relationships, and multi-modal cues. Compared to NeRF-based frameworks, 3DGS offers a more structured and interpretable formulation that enables efficient optimization, direct supervision, and intuitive manipulation. These properties make it advantageous for high-level tasks beyond rendering.

\begin{figure}[t]
    \centering
    \includegraphics[width=1.0\linewidth]{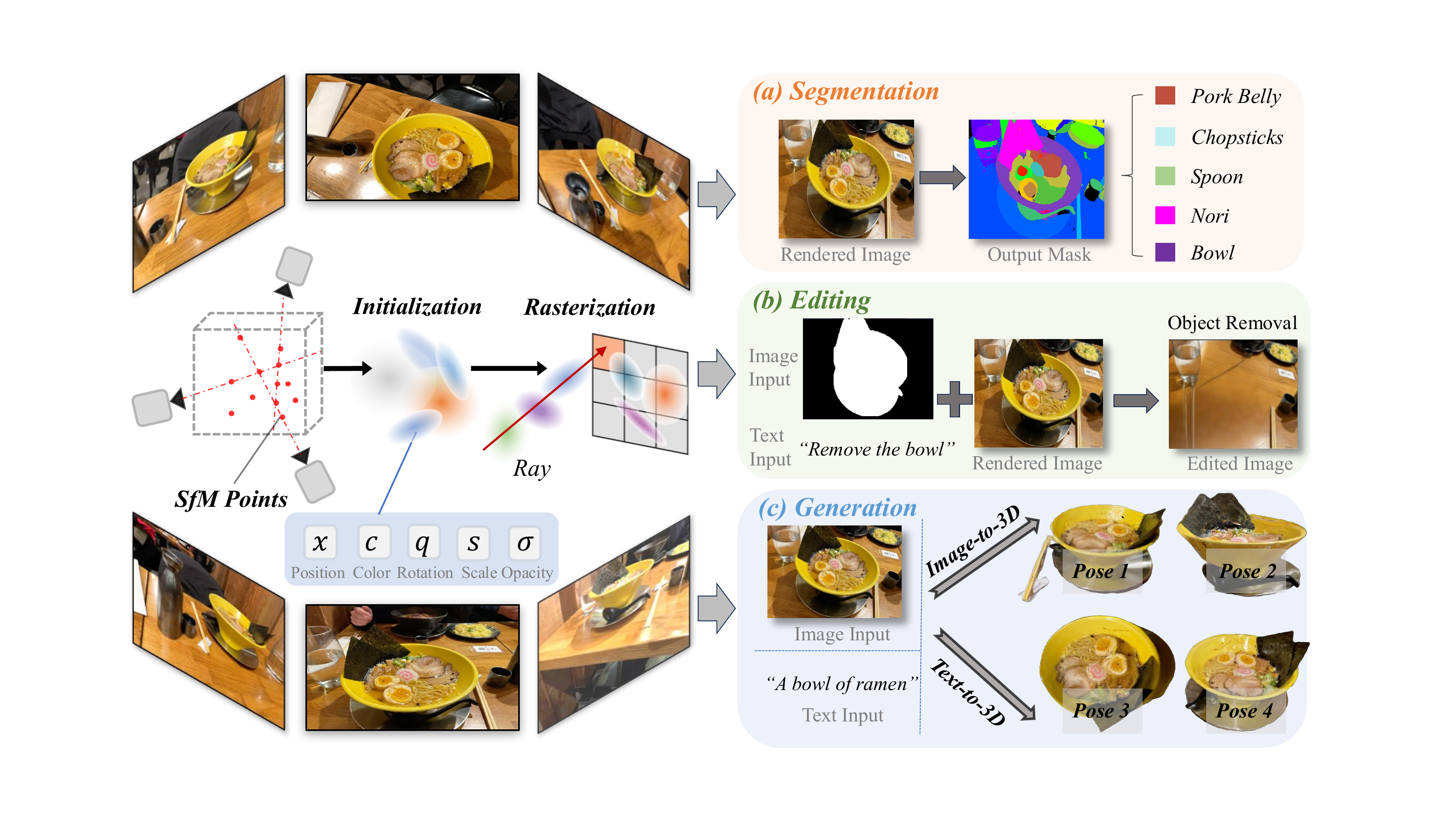}
    \vspace{-6mm}
    \caption{
        Overview of the three foundational 3DGS application tasks.
    (a) \textit{Segmentation}: Segment each semantic region.
    (b) \textit{Editing}: Perform editing via text or image prompt.
    (c) \textit{Generation}: Support text-to-3D and image-to-3D generation.
    }
	\vspace{-10pt}
     \label{fig:teaser}
\end{figure}

Although several recent surveys~\cite{chen2024survey, zhu20243d, bagdasarian20243dgs, ali2025compression} have documented the rapid development of 3DGS, they primarily focus on global taxonomies, real-time rendering pipelines, or compression strategies, while offering limited insights into 3DGS-driven downstream applications. A few works~\cite{wu2024recent, fei20243d, bao20253d} attempt to cover application domains, but typically lack systematic analysis of underlying design principles, methodological innovations, or benchmarking protocols. To address this gap, we present the first dedicated survey that systematically reviews downstream applications of 3DGS beyond classical view synthesis. 
Furthermore, we organize 3DGS application tasks into foundational and functional categories. The former represents general capabilities that are broadly used in various application scenarios, while the latter comprises scenario-driven pipelines tailored to specific applications built upon or tightly coupled with these capabilities. Accordingly, we focus primarily on three foundational tasks, as summarized in Fig.~\ref{fig:teaser}, and discuss functional tasks in the Appendix.

Although presented separately for clarity, these three tasks are closely interconnected. Segmentation provides semantic grounding and object-level decomposition, which are often prerequisites for precise and localized editing, such as modifying or removing a target object. Generation further extends this pipeline from modifying existing content to synthesizing new objects or scenes. Conversely, generative priors and synthetic data can in turn improve segmentation and editing quality, forming a mutually reinforcing loop. Rather than three parallel directions, these three tasks are deeply intertwined.


\begin{figure*}[t]
	\centering
	\includegraphics[width=1\linewidth]{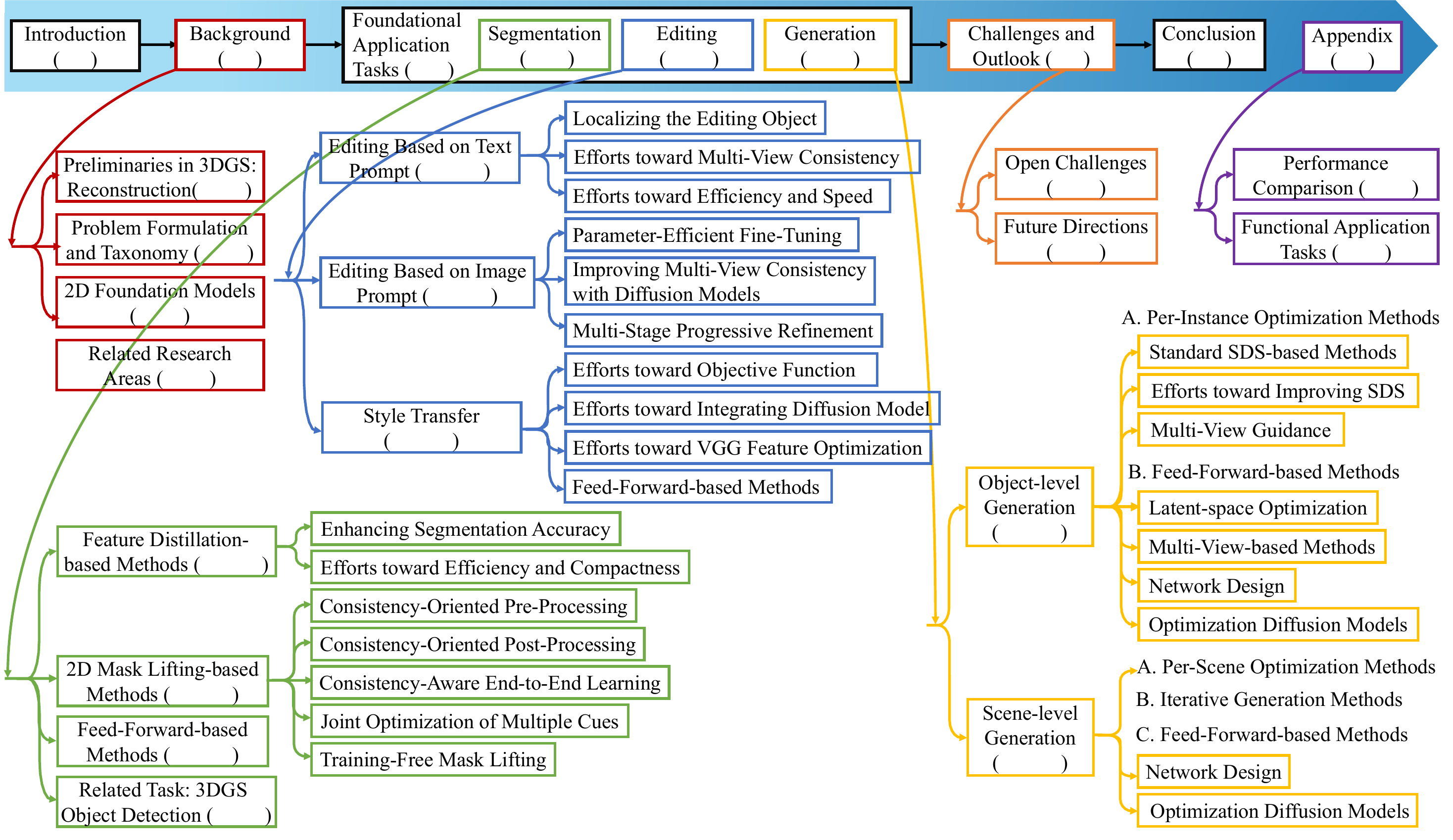}
        \put(-494.5,276){\footnotesize \S\ref{sec:introduction}}
	\put(-436,276){\footnotesize \S\ref{sec:background}}
	\put(-368,272){\footnotesize \S\ref{sec:method}}
	\put(-328.5,276){\scriptsize \S\ref{sec:3.1}}
	\put(-278,276){\scriptsize \S\ref{sec:3.2}}
    \put(-228,276){\scriptsize \S\ref{sec:3.3}}
	\put(-140,276){\footnotesize \S\ref{sec:challenge_outlook}}
    \put(-140,230){\scriptsize \S\ref{sec:challenges}}
	\put(-140,207.5){\scriptsize \S\ref{sec:future_direction}}
	\put(-90,276){\footnotesize \S\ref{sec:conclusion}}
    \put(-40,276){\scriptsize \S\textcolor{red}{A}}
    \put(-444,229){\scriptsize \S\ref{sec:bg_recon_nvs}}
	\put(-444,207){\scriptsize \S\ref{sec:2.1}}
	\put(-466.5,185){\scriptsize \S\ref{sec:2.2}}
	\put(-456.5,162.5){\scriptsize \S\ref{sec:2.3}}
	\put(-443,96){\middlefont \S\ref{sec:3.1.1}}
	\put(-453,50){\middlefont \S\ref{sec:3.1.2}}
	\put(-453,28){\middlefont \S\ref{sec:3.1.3}}
    \put(-440,6.5){\middlefont \S\ref{sec:object_detection}}
	\put(-364,236.5){\middlefont \S\ref{sec:3.2.1}}
	\put(-362,192){\middlefont \S\ref{sec:3.2.2}}
	\put(-375,140.5){\middlefont \S\ref{sec:3.2.3}}
    \put(-158,107.5){\middlefont \S\ref{sec:3.3.1}}
	\put(-158,25.5){\middlefont \S\ref{sec:3.3.2}}
    \put(-29.5,230 ){\scriptsize \S\ref{sec:performance}}
    \put(-39,207){\scriptsize \S\ref{sec:3.4}}
    \vspace{-6pt}
	\caption{Overview of the survey structure.}
	\vspace{-3pt}
	\label{fig:overview}
\end{figure*}

\noindent$\bullet$ \textbf{Contribution.} 
This survey provides a comprehensive and structured review of recent literature on 3DGS with a focus on its emerging downstream applications. 
We first review the reconstruction preliminaries of 3DGS, together with relevant 2D foundation models and related NeRF-based research areas, to establish conceptual continuity. 
The core of our survey is organized around three foundational task categories: segmentation, editing, and generation. Within each category, we systematically compare representative works in terms of technical designs, learning paradigms, and supervision strategies.
We further analyze benchmark settings and performance results in each domain, aiming to provide a self-contained and accessible resource for both newcomers and experienced researchers.
Finally, we discuss open challenges and future research opportunities, with the goal of fostering deeper exploration and broader adoption of 3DGS across high-level vision and graphics application tasks.

Fig.~\ref{fig:overview} provides an overview of the structure of this survey.
In Sec.~\ref{sec:background}, we review the essential background, including the reconstruction preliminaries of 3DGS, the problem formulation and taxonomy of downstream applications, commonly used 2D foundation models, and related NeRF-based research areas that inform 3DGS development.
Sec.~\ref{sec:method} presents a task-centric review of recent 3DGS-based methods, with a focus on segmentation, editing, and generation.
Sec.~\ref{sec:challenge_outlook} discusses open challenges and outlines potential future research directions.
Sec.~\ref{sec:conclusion} concludes the survey with a summary of key insights and takeaways.
Finally, we provide additional discussions of performance comparisons among recent methods and functional application tasks in the Appendix.

\section{Background}
\label{sec:background}
In this section, we first review the preliminaries of 3DGS-based reconstruction, and then outline the main research directions for downstream applications. We further provide a concise overview of widely used 2D foundation models that support these tasks, and finally discuss related research domains in the NeRF literature that have laid the foundation for the development of 3DGS.

\subsection{Preliminaries in 3DGS: Reconstruction}\label{sec:bg_recon_nvs}

Although this survey focuses on downstream application tasks such as segmentation, editing, and generation, reconstruction and novel view synthesis (NVS) remain the historical and technical foundation of the broader 3DGS ecosystem. Revisiting these reconstruction- and representation-oriented foundations helps clarify the technical basis upon which downstream applications are built. Below, we briefly summarize several major upstream directions that provide important context for, yet remain complementary to, the main focus of this survey.

\noindent~$\bullet$~\textbf{Vanilla 3DGS Reconstruction.}
The original 3D Gaussian Splatting (3DGS) framework~\cite{3DGS} is introduced for high-fidelity 3D reconstruction and real-time novel-view synthesis. It represents a scene as a set of explicit 3D Gaussian primitives~\( \mathcal{G} = \{ g_i \}_{i=1}^{\mathcal{N}} \), where \( \mathcal{N} \) is the number of Gaussians. Each Gaussian \( g_i \) is parameterized by its {mean position} \( \mu_i \in \mathbb{R}^3 \), {covariance matrix} \( \Sigma_i \in \mathbb{R}^{3 \times 3} \), {opacity} \( \sigma_i \in \mathbb{R}^1 \), and {color} \( c_i \in \mathbb{R}^{d_c} \), where \( d_c = 3 \) for RGB color parameters.
To render an image, each 3D Gaussian is projected onto the 2D image plane, where it contributes to the pixel color based on its opacity. The color at a given pixel \( v \), denoted as \( C(v) \), is computed by blending the contributions of all Gaussians according to their opacity. This process is formulated as:
\vspace{-1mm}\begin{equation}
C(v) = \sum_{i=1}^{\mathcal{N}} c_i \alpha_i \prod_{j=1}^{i-1} (1 - \alpha_j),
\end{equation}
where \( c_i \) is the color vector of the \( i \)-th Gaussian, and \( \alpha_i = \sigma_i G^{2D}_i(v) \), \( G^{2D}_i(v) \) is the 2D projection function. 
This explicit representation renders substantially faster than NeRF-style implicit pipelines while preserving strong reconstruction quality. However, vanilla 3DGS is mainly tailored to dense-view, static, and well-posed scenes, motivating many subsequent extensions.

\noindent~$\bullet$~\textbf{Sparse-view Reconstruction.}
An important line of research studies how to extend 3DGS from dense-view settings to sparse-view or few-shot reconstruction, where limited observations often lead to geometric degradation, overfitting to seen views, and unstable novel-view synthesis~\cite{xiong2023sparsegs,zhu2023fsgs,li2024dngaussian}. Existing methods mainly follow two directions. The first is {regularization-based reconstruction}, which introduces additional geometric cues such as monocular or multi-view depth priors, smoothness constraints, cost-volume guidance, or Gaussian densification strategies to improve surface completeness and global consistency under sparse inputs~\cite{chung2023depth,zhu2023fsgs,xiong2023sparsegs,li2024dngaussian,chen2024mvsplat,zhang2024cor}. The second direction leverages stronger learned priors, either by directly inferring Gaussian representations from sparse observations or by augmenting sparse inputs with generative models~\cite{pixelSplat,szymanowicz2023splatter,xu2024grm,szymanowicz2024flash3d,sargent2024zeronvs}.

\noindent~$\bullet$~\textbf{Generalizable 3DGS.}
To overcome the limitation of per-scene optimization in vanilla 3DGS, a growing body of work studies {generalizable 3DGS}, which aims to predict Gaussian representations for unseen scenes directly from one or multiple input images using learned priors. Existing methods can be broadly grouped by the form of the predicted representation. One line predicts {Gaussian maps}, where each image location encodes a 3D Gaussian, typically using encoder-decoder or Transformer-based architectures for pixel-aligned Gaussian prediction~\cite{szymanowicz2024splatter,xu2024grm,GS-LRM,min2024epipolar,ziwen2024long,xu2024freesplatter}. To improve geometric quality, many methods further incorporate multi-view geometric reasoning, such as epipolar constraints and cost-volume representations, for more reliable depth and cross-view correspondence estimation~\cite{pixelSplat,wewer2024latentsplat,li2024ggrt,chen2024mvsplat,xu2024depthsplat,zhang2025transplat}. Another line predicts more structured {Gaussian volumes} or triplane-based representations, which provide stronger 3D inductive bias for feed-forward reconstruction~\cite{chen2024lara,zou2024triplane,xu2024agg,zhang2024gaussiancube,liu2025quicksplat,jiang2025anysplat}. In addition, several methods leverage pre-trained reconstruction or generative foundation models as geometry or appearance priors, further improving generalization under sparse-view or pose-free settings~\cite{smart2024splatt3r,ye2024no,kang2024selfsplat,tang2024lgm,liang2025wonderland}.

\noindent~$\bullet$~\textbf{Efficiency and Compression.}
Although 3DGS enables high-quality real-time rendering, its practical deployment still faces considerable memory and storage challenges due to the large number of Gaussians and their associated attributes. Existing efficiency-oriented studies mainly improve compactness from two perspectives. The first is to {reduce the number of Gaussians} through pruning, importance-based selection, or more structured representations such as anchor-, grid-, or octree-based designs~\cite{Scaffold-gs,lee2023compact,fan2024lightgaussian,chen2024hac,ren2024octree,papantonakis2024reducing,fang2024mini}. The second is to {compress Gaussian attributes} by using vector quantization, codebooks, residual quantization, or other compact storage schemes for color, spherical harmonics, scale, rotation, and related properties~\cite{niedermayr2024compressed,lee2023compact,girish2023eagles,fan2024lightgaussian,navaneet2023compact3d}. Some methods further combine both strategies by coupling structured representations with adaptive quantization to reduce redundancy while maintaining rendering quality~\cite{chen2024hac,Scaffold-gs}.

\noindent~$\bullet$~\textbf{Dynamic Reconstruction.}
Beyond static scene reconstruction, 3DGS has also been extended to dynamic 3D reconstruction, where the core challenge is to model time-varying scene content while preserving geometric fidelity and temporal consistency. In general, dynamic reconstruction methods typically need to address two key issues: how to separate static and dynamic components in the scene, and how to represent the motion of dynamic objects. Existing methods mainly follow three paradigms. The first is {time-varying modeling}, which directly parameterizes Gaussian attributes such as position, rotation, opacity, or appearance as functions of time~\cite{luiten2023dynamic,chen2023periodic,li2024vdg,li2023spacetime,zhou2023drivinggaussian,yan2024street}. The second is {deformation-based modeling}, which defines a canonical static space and learns a deformation field to warp Gaussians across time~\cite{yang2023deformable,wu20234d,lin2023gaussian,sun20243dgstream,lu20243d,fischer2024dynamic}. The third is {4D Gaussian modeling}, which treats spacetime as a unified entity by extending 3D Gaussians into 4D primitives~\cite{yang2023real,duan20244d}. These methods have enabled dynamic reconstruction in a wide range of scenarios, from object- and human-centric settings to large-scale driving and urban environments~\cite{lin2024vastgaussian,zhou2023drivinggaussian,chen2023periodic}.

\noindent~$\bullet$~\textbf{Mesh Extraction and Surface Recovery.}
Beyond view synthesis, reconstruction-oriented 3DGS research has also explored mesh extraction and explicit surface recovery, which are important for applications requiring explicit geometry, physical interaction, or geometry-aware editing. Existing methods typically combine Gaussian representations with surface regularization or hybrid geometric formulations to improve the recoverability of explicit surfaces~\cite{guedon2023sugar,chen2023neusg}. In this setting, evaluation involves not only rendering fidelity but also geometric quality criteria such as surface accuracy, completeness, smoothness, and topology preservation.

\noindent~$\bullet$~\textbf{Photorealism and Quality Enhancement.}
Beyond efficient novel view synthesis, another important research direction aims to improve the visual fidelity of 3DGS rendering. Existing studies mainly enhance photorealism from three aspects. The first is to reduce {aliasing and splatting artifacts}, which often arise from the discrete sampling and projection process of Gaussian splatting, especially under varying resolutions or focal lengths~\cite{yan2023multi,yu2023mip,zhang2024fregs,liang2024analytic}. The second is to improve {view-dependent appearance}, including reflections, specularities, and relightable effects, by incorporating more expressive lighting and shading models, inverse rendering formulations, or physically inspired appearance representations~\cite{gao2023relightable,jiang2023gaussianshader,liang2023gs,ma2023specnerf,meng2024mirror,yang2024spec}. The third is to address {real-world degradations}, such as motion blur, defocus blur, and unstable visibility, which can significantly reduce rendering realism in practical captures~\cite{peng2024bags,zhao2024bad,radl2024stopthepop}.

\noindent~$\bullet$~\textbf{Physically Grounded 3D Modeling.}
Beyond geometrically plausible reconstruction and rendering, an emerging direction aims to make 3DGS physically grounded, so that dynamic scene evolution is not only visually convincing but also consistent with underlying physical behaviors. Existing works typically build on dynamic or 4D Gaussian representations~\cite{wu20234d,yang2023real,lu20243d,duisterhof2023md,kratimenos2023dynmf,sun20243dgstream} and further introduce physics-aware mechanisms in two main ways. One line incorporates explicit simulation priors, such as material point methods, position-based dynamics, or other mechanics-inspired simulators, by treating Gaussians as particles or material carriers for deformation and motion~\cite{xie2024physgaussian,liu2024physics3d,borycki2024gasp,qiu2024feature,huang2024dreamphysics,zhang2024physdreamer,jiang2024vr,feng2024gaussian,zhong2024reconstruction}. Another line imposes physically motivated constraints on motion and deformation, such as rigidity, isometry, momentum conservation, to encourage more faithful and temporally consistent dynamics~\cite{duisterhof2023md,kratimenos2023dynmf}.

\subsection{Problem Formulation and Taxonomy}\label{sec:2.1}

Let \bm{$\mathcal{X}$} and \bm{$\mathcal{Y}$} denote the input and output spaces, respectively. In the context of 3DGS-based downstream applications, the objective is to learn an ideal mapping function $f^{*!}:\bm{\mathcal{X}}\mapsto\bm{\mathcal{Y}}$ that transforms visual observations into task-specific outputs such as semantic understanding, controllable editing, or content generation. Specifically, given a set of posed RGB images \(\{x_n^s\}_{n,s} \subset \bm{\mathcal{X}}\), where \( n \) denotes the number of views, $s$ represents different scenes, 3DGS provides a differentiable intermediate representation that bridges perception and application-level tasks.

\subsubsection{Application Tasks Category}\label{sec:2.1.1}
Downstream applications built upon 3DGS can be broadly categorized into {foundational} and {functional} tasks. In this survey, we focus on three representative foundational tasks: segmentation, editing, and generation, which differ in their output space $\bm{\mathcal{Y}}$.

\noindent$\bullet$~\textbf{3DGS Segmentation.}
Based on the input space \bm{$\mathcal{X}$}, existing 3DGS segmentation methods can be broadly classified into two categories: generic segmentation and promptable segmentation.
Generic segmentation aims to predict pixel- or point-level labels without external input prompts, and can be further divided into:
(i) Semantic segmentation~\cite{RT-GS2};
(ii) Instance segmentation~\cite{OmniSeg3D,Gaga,Unified-Lift}; and
(iii) Panoptic segmentation~\cite{CCGS,GS-Grouping}.
Promptable segmentation extends the generic setting by introducing external input prompts to guide the segmentation process. Depending on the form of prompts, this category includes:
(i) Interactive segmentation, where users provide spatial hints such as clicks, scribbles, boxes, or masks~\cite{SAGD,Click-Gaussian,iSegMan};
(ii) Open-vocabulary segmentation, which uses category names as textual prompts to support segmenting novel or unseen classes~\cite{LangSplat,Feature-3DGS};
(iii) Referring segmentation, which relies on natural language expressions to localize and segment specific objects~\cite{Ref-LERF}.
In addition, depending on the representation of the output space \bm{$\mathcal{Y}$}, segmentation results can be produced either on rendered 2D images~\cite{LangSplat,3DVision-LanguageGS} or directly over the 3D Gaussian representations~\cite{GaussianCut,InstanceGaussian,OpenGaussian}.

\noindent$\bullet$~\textbf{3DGS Editing.}
Editing tasks aim to modify scene attributes such as geometry, appearance, or illumination, while maintaining structural and visual consistency across viewpoints. In this setting, \bm{$\mathcal{Y}$} denotes the edited scene state, often represented by an updated set of Gaussians. 3DGS facilitates localized and differentiable editing operations, including addition, deletion, and transformation of scene components, thus enabling applications such as content-aware inpainting and style-consistent relighting. Based on the form of inputs \bm{$\mathcal{X}$}, editing approaches can be categorized into image-guided~\cite{TIP-Editor,vcEdit} and text-guided methods~\cite{GaussianEditor-HGS,GaussianEditor-ROI}.

\noindent$\bullet$~\textbf{3DGS Generation.}
Generation involves synthesizing novel 3D scenes or objects from limited inputs such as a single image, a sparse set of views, or a textual prompt. Here, \bm{$\mathcal{Y}$} represents the desired 3D output in the form of Gaussian primitives. Recent approaches enable direct generation of structured 3D content, supporting high-quality synthesis and compositionality within multimodal AIGC frameworks. Depending on the modality of inputs, generation methods can be grouped into image-to-3D~\cite{LGM,GRM,GS-LRM} and text-to-3D paradigms~\cite{yi2024gaussiandreamer,liang2024luciddreamer,hollein2023text2room}.
Alternatively, generation methods can be categorized by output granularity into object-level~\cite{Dreamgaussian,yi2024gaussiandreamer} and scene-level generation~\cite{DreamScene,DreamScene360}.

\subsubsection{Learning Paradigms for Application Tasks}
\noindent$\bullet$~\textbf{Per-scene Optimization.}  
In this paradigm, the model learns a scene-specific mapping \( f_{\theta_s} \) for each individual scene \( s \), where \(\theta_s\) denotes parameters optimized solely for that scene. Given multi-view observations \(\{x_n^s\}_n\) for one scene, the objective is to minimize a reconstruction loss across views:
\begin{equation}
\tilde{\theta}_s \in \mathop{\arg\min}_{\theta_s} \frac{1}{N} \sum\nolimits_n \varepsilon(f_{\theta_s}(x_n^s), x_n^s),
\end{equation}
where \( \varepsilon(\cdot, \cdot) \) denotes a task-specific reconstruction loss function, such as an \(\ell_1\) loss, depending on the target application.
This per-scene learning scheme enables high-fidelity modeling tailored to each scene, making it well-suited for tasks requiring view-consistent reconstruction or dense appearance modeling. However, the lack of generalization and high computational cost make it impractical for large-scale or real-time applications.

\noindent$\bullet$~\textbf{Feed-forward Learning.}  
This alternative seeks to approximate a universal function \( f_\theta \) that generalizes across scenes. A shared parameter set \(\theta\) is learned from a collection of scenes \(\{x_n^s\}_{n,s} \subset \bm{\mathcal{X}}\) drawn from diverse environments:
\begin{equation}
\tilde{\theta} \in \mathop{\arg\min}_{\theta} \frac{1}{NS} \sum\nolimits_{n,s} \varepsilon(f_\theta(x_n^s), x_n^s).
\end{equation}
The learned function \( f_\theta \) allows direct inference on novel scenes via a single forward pass, supporting efficient generation and scalable deployment. While typically less precise than per-scene methods in modeling fine-grained geometry or photometric details, feed-forward approaches offer strong potential for real-time applications and serve as a foundation for generalizable 3D perception.

\subsection{2D Foundation Models}\label{sec:2.2}
The success of 3DGS applications relies heavily on the integration of powerful 2D foundation models. Given the limited availability of large-scale 3D datasets, learning robust 3D representations remains challenging. To address this, a number of methods leverage pretrained 2D priors to enhance 3D representation.

\noindent~$\bullet$~\textbf{{Foundation Features for Vision Tasks (DINO\&DINOv2).}} DINO \cite{DINO} frames self-supervised ViT training as self-distillation without labels, where the model learns from its own predictions. Despite its moderate size, it yields semantically rich features effective for object understanding. DINOv2~\cite{oquab2023dinov2} scales this approach with larger models, more data, and refined training, producing strong general-purpose features competitive with supervised baselines. These models reveal emergent capabilities in part-level reasoning and robust representations, making them valuable.

\noindent~$\bullet$~\textbf{{Contrastive Language-Image Pre-training (CLIP).}} 
CLIP \cite{CLIP} trains separate image and text encoders on 400M web pairs, using a contrastive loss that brings matched embeddings together in a shared space while pushing mismatches apart.
The resulting representation carries broad semantic knowledge and strong zero-shot ability, making CLIP a standard backbone for open-vocabulary and multimodal tasks.

\noindent~$\bullet$~\textbf{{Segment Anything (SAM)}}. SAM \cite{SAM} is a foundation model for image segmentation, trained on 1B masks over 11M images using a promptable paradigm. It supports versatile inputs—points, boxes, masks, or text—and enables zero-shot generalization across diverse segmentation tasks. While powerful, its heavy vision backbone limits efficiency, SAM2~\cite{ravi2024sam} mitigates this with a hiera-based encoder, achieving real-time performance and improved accuracy in the video domain~\cite{MOSEv2,MOSE,MeViS,MeViSv2}.

\noindent~$\bullet$~\textbf{{Diffusion Models (DMs)}}.
These generative models start from Gaussian noise and progressively denoise it into an image (DDPM~\cite{ho2020denoising}). Latent Diffusion Models~\cite{rombach2022high} (LDM) improve computational efficiency by performing denoising in a compressed latent space, while Stable Diffusion~\cite{rombach2022high} builds on this approach by enabling text-conditioned generation through a CLIP~\cite{CLIP} text encoder and a cross-attention mechanism. DiT~\cite{peebles2023scalable} replaces the U-Net backbone with a Transformer, enhancing scalability and performance, and Flux~\cite{flux2024} further advances this line by adopting a flow-matching framework with Transformer backbones, enabling faster sampling and higher-fidelity image generation.

\subsection{Related Research Areas}\label{sec:2.3}

\noindent$\bullet$ \textbf{Segmentation Based on NeRF.}
Semantic NeRF~\cite{zhi2021place} pioneers the incorporation of semantic information into radiance fields by introducing per-point semantic layers. Follow-up works can be broadly categorized into two directions. The first focuses on feature distillation, where high-dimensional 2D features from pretrained vision models (\eg, DINO, CLIP) are embedded into 3D neural fields. Representative methods such as Distilled Feature Fields~\cite{kobayashi2022decomposing}, Neural Feature Fusion Fields~\cite{tschernezki2022neural}, ISRF~\cite{ISRF}, LERF~\cite{LERF}, and 3D-OVS~\cite{3D-OVS} optimize volumetric feature fields to reconstruct semantic features via differentiable rendering.
The second line of research targets mask lifting, which seeks to overcome the scarcity of 3D annotations by projecting 2D semantic masks into 3D space. Early methods rely on either ground-truth masks~\cite{wang2022dm} or pretrained segmentation models, such as methods like Panoptic Lifting~\cite{PanopticLifting}, ContrastiveLift~\cite{ContrastiveLift}, and SPIn-NeRF~\cite{SPIn-NeRF}. However, these approaches often suffer from limited generalization due to closed-set assumptions.
With the recent emergence of powerful foundation models like SAM and SAM2~\cite{SAM}, newer works aim to leverage their open-world segmentation capabilities. For instance, SA3D~\cite{SA3D} proposes an interactive pipeline that propagates a single SAM mask across multiple views into a NeRF-based scene, enabling sparse-to-dense segmentation. 
GARField~\cite{kim2024garfield} integrates SAM-derived masks into radiance field by learning a scale-conditioned affinity field.

\noindent$\bullet$ \textbf{Editing Based on NeRF.}
Recent progress in NeRF-based editing and manipulation has led to more intuitive and controllable frameworks for modifying 3D scenes and objects. Early efforts introduce conditional radiance fields that support localized editing through sparse 2D user inputs, enabling view-consistent changes in appearance and shape with minimal supervision~\cite{liu2021editing}. Building on this, CLIP-NeRF~\cite{Clip-nerf} pioneers text- and image-driven NeRF editing using disentangled latent spaces for geometry and appearance, allowing independent control guided by CLIP embeddings and supporting both category-level generation and real image inversion. 
Subsequent methods have addressed object-level editing and inpainting. 
Object-NeRF~\cite{Object-NeRF} and LaTeRF~\cite{Laterf} further explore object-centric manipulation and structural completion in NeRF scenes. CoNeRF~\cite{Conerf} introduces fine-grained, attribute-specific control with few-shot supervision, leveraging sparse 2D mask annotations and a quasi-conditional latent structure to enable localized edits and novel expression synthesis. Unified models that combine segmentation with NeRF-based inpainting further enable intuitive scene manipulation, such as object removal and completion from sparse cues~\cite{SPIn-NeRF}. Instruct-NeRF2NeRF~\cite{haque2023instructnerf2nerfediting3dscenes} employs an image-conditioned diffusion model to iteratively edit input images while jointly optimizing the underlying 3D scene.

\begin{figure}[t]
    \centering
    \includegraphics[width=1\linewidth]{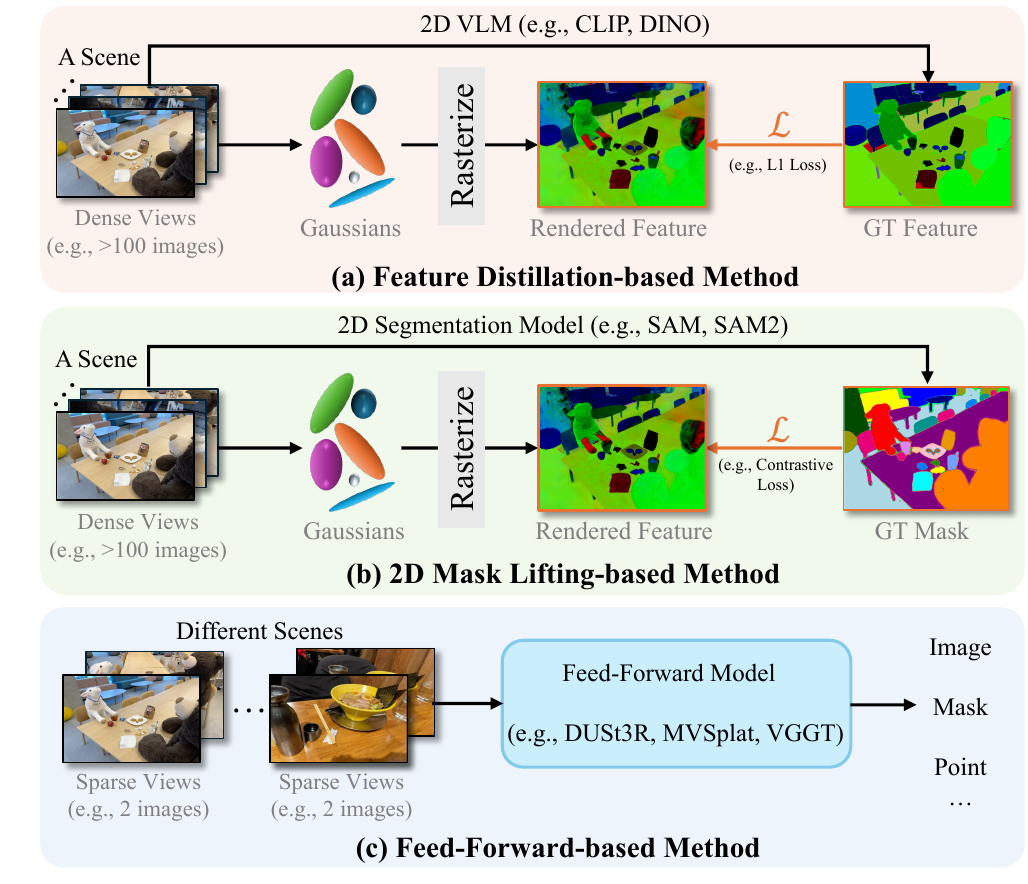}
    \vspace{-3mm}
    \caption{A comparison of three 3DGS segmentation pipelines.
    }
    \label{fig:seg_task}
\vspace{-6pt}
\end{figure}

\noindent$\bullet$ \textbf{Generation Based on NeRF.}
NeRF-based generative models have rapidly advanced 3D-aware content synthesis. Early works utilize implicit volumetric representations with adversarial training to achieve 3D consistent image generation from unposed views, improving shape–appearance disentanglement and scalability \cite{schwarz2020graf}. To boost efficiency and detail, later methods constrain sampling to 2D manifolds via ray-surface intersections \cite{deng2022gram}. A major shift occurs with Score Distillation Sampling (SDS), which enables NeRF optimization from text prompts using 2D diffusion models, eliminating the need for 3D supervision \cite{poole2022dreamfusion}. Follow-up work adopts coarse-to-fine strategies and hybrid representations to enhance quality and speed \cite{lin2023magic3d}. Latent-NeRF~\cite{metzer2023latent} uses Sketch-Shape to define the coarse structure of abstract geometric objects, enabling text- and shape-guided control over the generation process. Wang~\etal~\cite{wang2023score} apply the chain rule to the learned gradients and backpropagate the diffusion model's score through the Jacobian of a differentiable renderer, instantiated as a voxel radiance field. In parallel, GAN-based models introduce tri-plane features and pose-aware training to generate high-resolution, multi-view consistent outputs \cite{chan2022efficient}. Recent diffusion-guided approaches further extend to full-scene generation by combining semantic priors~\cite{zhang2024text2nerf}.

\vspace{-1.2mm}
\section{Foundational Application Tasks}
\label{sec:method}

\subsection{3DGS Segmentation}\label{sec:3.1}

In 3DGS-based scene understanding, segmentation and object detection are closely related tasks and often share similar representations and learning paradigms. 
We begin with segmentation, as it has been more extensively studied and provides a natural basis for organizing related scene understanding techniques.
For completeness, we then review object detection in Sec.~\ref{sec:object_detection} as a closely related task to segmentation, emphasizing both shared components and task-specific differences.
Segmentation encompasses a broad spectrum of task definitions, as discussed in Sec.~\ref{sec:2.1.1}. 
Given the substantial overlap among task settings, where a single method may apply to multiple scenarios, we organize existing approaches by their technical characteristics rather than by task formulations. 
Representative segmentation pipelines are provided in Fig.~\ref{fig:seg_task}.

\subsubsection{Feature Distillation-based Methods}\label{sec:3.1.1}
These methods~\cite{LangSplat,Feature-3DGS,LEGaussians,MaskField,SemanticGaussians} aim to distill the semantic knowledge embedded in 2D foundation models (\eg, CLIP\cite{CLIP}, SAM~\cite{SAM}, DINO~\cite{DINO}) into 3D scene representations. By leveraging the semantic understanding learned from large-scale 2D datasets, these approaches enhance 3D models with open-vocabulary recognition capabilities.
LangSplat~\cite{LangSplat} and Feature3DGS~\cite{Feature-3DGS} are among the earliest works to explore this direction.
LangSplat first segments images using SAM~\cite{SAM}, and then feeds the resulting hierarchical semantic masks into CLIP to extract region-level semantic embeddings. To alleviate memory constraints, it employs a scene-specific language autoencoder to compress these high-dimensional semantic features.
These compressed features are used to supervise a set of initialized 3D Gaussian language features, effectively transferring CLIP's open-vocabulary semantics into the 3D domain for downstream tasks.
Feature3DGS, developed concurrently, distills features from LSeg~\cite{Lseg} and SAM~\cite{SAM}, and further employs SAM’s decoder to interpret 2D rendered views, establishing a bridge between 2D and 3D segmentation.
While effective, these methods may suffer from semantic information loss and high computational cost. To address these limitations, subsequent works have evolved along two key directions: improving segmentation accuracy and reducing computation overhead.

\noindent $\bullet$ \textbf{Enhancing Segmentation Accuracy.}
N2F2~\cite{N2F2} addresses LangSplat’s inefficiency in scale selection by learning a unified high-dimensional feature field under hierarchical supervision, where each feature dimension encodes scene semantics at different granularities, enabling fine-grained yet efficient representation without multi-scale querying.
LangSurf~\cite{LangSurf} derives dense pixel-level semantics with SAM pooling, and jointly optimizes surface language Gaussians with geometric supervision and contrastive loss for improved 3D spatial consistency.
seconGS~\cite{seconGS} refines LangSplat with SAM2 masklet supervision and a two-step querying scheme that first retrieves distilled ground-truth and then queries individual Gaussians for more accurate 3D understanding.
VLGaussian~\cite{3DVision-LanguageGS} observes that prior color-based rasterization is unsuitable for language modalities due to its dependence on the same color opacity. Instead, it introduces a cross-modal rasterizer tailored for rendering semantic language features with newly introduced opacity for language.
SuperGSeg~\cite{SuperGSeg} clusters Gaussians into Super-Gaussians with spatial and instance cues built on Scaffold-GS~\cite{Scaffold-gs} and assigns language features per cluster for structured scene understanding.
FreeGS~\cite{FreeGS} introduces a semantic-embedded 3DGS framework by coupling semantic features with instance identities via the proposed IDSF field. A two-step alternating optimization and a 2D–3D contrastive loss ensure view-consistent semantics without requiring 2D labels.
econSG~\cite{econSG} improves zero-shot 3D semantic understanding by projecting refined multi-view language features into a low-dimensional 3D latent space, enabling efficient optimization and consistent semantic field initialization within the 3D Gaussian framework.
CCL-LGS~\cite{CCL-LGS} aligns SAM-generated masks across views via a zero-shot tracker, extracts semantics with CLIP, and applies contrastive codebook learning to enforce cross-view consistency by promoting intra-class compactness and inter-class separation.
ILGS~\cite{ILGS} mitigates view-inconsistent language embeddings by enforcing cross-view semantic alignment with an identity-aware semantic consistency loss and refining boundaries via progressive mask expansion.
COS3D~\cite{COS3D} learns collaborative instance and language fields, using SAM masks with contrastive instance supervision and an instance-to-language mapping. At inference, it uses adaptive language-to-instance prompt refinement to produce accurate output.
Beyond the aforementioned approaches that solely introduce semantic information, GLS~\cite{GLS} incorporates geometric priors (surface normals) to effectively refine the rendered depth.
REALM~\cite{REALM} enables reasoning-based open-world 3D segmentation via a global-to-local MLLM grounding strategy, avoiding extensive 3D-specific post-training while supporting downstream 3D interaction tasks.

\noindent $\bullet$ \textbf{Efforts toward Efficiency and Compactness.}
To mitigate the substantial memory overhead incurred by increasing feature channels in raw semantic features, LEGaussian~\cite{LEGaussians} quantizes dense language features into a discrete feature space and semantic indices, and applies a smoothing loss to retain rendering quality with quantized representation.
CLIP-GS~\cite{CLIP-GS} introduces semantic attribute compactness to efficiently encode scene semantics with Gaussians, enabling fast training and inference. It further improves 3D segmentation by employing a 3D-coherent self-training strategy that enhances cross-view semantic consistency.
Later, FMGS~\cite{FMGS} integrates 3D Gaussian scene representation with multi-resolution hash encodings (MHE) to enable efficient semantic embedding. Unlike previous methods, it avoids scene-specific quantization or autodecoders, thereby preserving the semantic fidelity of foundation model features.
GOI~\cite{GOI} learns a clustering codebook to compress noisy high-dimensional semantic features into compact, scene-prior-guided representations, reducing storage and rendering cost.
FastLGS~\cite{FastLGS} constructs a semantic feature grid that stores multi-view CLIP features extracted from SAM-generated masks. These grid features are then projected into a low-dimensional space and used to supervise semantic field training.
Building upon FastLGS, FMLGS~\cite{FMLGS} enables part-level open-vocabulary queries in 3DGS by constructing consistent object-part semantics via SAM2, and introduces a semantic deviation strategy to enrich fine-grained features, allowing natural language queries over both objects and their parts.
To address the high memory cost of jointly encoding color and semantics in existing 3DGS-based methods\cite{Feature-3DGS,LEGaussians,CLIP-GS}, DF-3DGS~\cite{DF_3DGS} decouples color and semantic fields to reduce Gaussian usage for semantics, and introduces a hierarchical compression pipeline combining dynamic quantization and scene-specific autoencoding for efficient representation.
%
LansSplatV2~\cite{LangSplatV2} represents each Gaussian’s CLIP feature using a global codebook with sparse top-$K$ coefficients, rendering only the coefficients for efficient open-vocabulary querying.
In summary, knowledge distillation provides a practical mechanism to bridge 2D foundation models and 3D Gaussian representations. While early methods focus on direct supervision, recent advances introduce architectural, supervisory, and efficiency-oriented innovations to improve precision and scalability. These works collectively advance open-vocabulary 3D understanding under constrained computational and data resources.

\subsubsection{2D Mask Lifting-based Methods}\label{sec:3.1.2}
Accurate 3DGS segmentation remains challenging due to limited annotation and costly labeling. To address this, recent works explore lifting 2D masks from foundation models (\eg, SAM\cite{SAM}, SAM2~\cite{ravi2024sam}) into 3D. However, 2D masks often suffer from cross-view inconsistency, where the same object may receive different instance IDs from different views, as well as quality issues like over- and under-segmentation, making direct lifting non-trivial.

\noindent $\bullet$ \textbf{Consistency-Oriented Pre-Processing.}
Several methods~\cite{GS-Grouping,Gaga,CCGS,GradiSeg,Cosseggaussians,segmentsplat} aim to improve multi-view mask consistency through pre-processing strategies before lifting 2D masks into 3D.
GaussianGrouping~\cite{GS-Grouping} introduces an object association technique as a pre-processing step to align 2D segmentation maps across views, enhancing cross-view consistency. However, the quality of such pre-processing is often limited by errors in object correspondence under varying viewpoints.
Gaga~\cite{Gaga} addresses this by leveraging spatial cues and a 3D-aware memory bank to effectively associate object masks across diverse camera poses.
CCGS~\cite{CCGS} further argues that Gaga lacks segmentation-aware optimization within the Gaussian representation itself, which may lead to inconsistent or floating structures in the 3D segmentation field. To overcome this, CCGS constructs a unified 3D point cloud field and performs segmentation-constrained optimization to reinforce mask association.

\noindent $\bullet$ \textbf{Consistency-Oriented Post-Processing.}
While pre-processing strategies help improve mask alignment across views, they are often susceptible to accumulated errors that degrade segmentation quality. To address this, recent methods~\cite{OmniSeg3D,CAGS} have shifted focus to post-processing techniques that refine multi-view predictions after initial lifting.
For instance, OmniSeg3D~\cite{OmniSeg3D} encodes instance-level semantics into a feature field through hierarchical contrastive learning. It then applies a hierarchical clustering algorithm in post-processing to merge and refine the lifted masks, ultimately yielding consistent and coherent 3D instance segmentations across views.
Similarly, CAGS~\cite{CAGS} leverages instance clustering and semantic matching to associate 3D instances with 2D open-vocabulary priors, thereby enhancing semantic consistency.

\noindent $\bullet$ \textbf{Consistency-Aware End-to-End Learning.}
To reduce reliance on heuristic pre- and post-processing, recent works~\cite{SAGA,SA3D-GS,Click-Gaussian,OpenGaussian,CGC,RT-GS2,OpenSplat3D} propose end-to-end learning frameworks that directly construct consistent and distinguishable 3D feature fields.
SAGA~\cite{SAGA} distills 2D masks into 3D features that introduce a scale-aware contrastive training strategy that distills SAM's segmentation capability into a scale-gated affinity representation, which handles multi-granularity ambiguity.
SA3D~\cite{SA3D-GS} introduces an iterative refinement strategy that alternates between inverse rendering of 2D masks and cross-view self-prompting, enabling progressively more consistent and accurate object segmentation across multiple views.
Click-Gaussian~\cite{Click-Gaussian} enables efficient interactive segmentation of 3D Gaussians by incorporating multi-granularity feature fields derived from 2D masks. It further introduces Global Feature-guided Learning (GFL) to alleviate inconsistencies across different views.
Unified-Lift~\cite{Unified-Lift} develops an object-aware lifting pipeline that eliminates the need for additional alignment steps. It formulates a codebook-based object representation and aligns object-level semantics with Gaussian features via contrastive learning, leading to improved 3D instance understanding.
OpenGaussian~\cite{OpenGaussian} focuses on 3D open-vocabulary segmentation by leveraging SAM-predicted masks to supervise 3D instance features. It adopts a coarse-to-fine feature discretization strategy via a two-stage codebook and introduces an instance-level 3D–2D association module that bridges Gaussian points with 2D masks and CLIP embeddings.
VoteSplat~\cite{VoteSplat} integrates Hough voting with 3DGS by learning per‑Gaussian 3D offsets whose projections align with SAM‑derived 2D vote maps and are depth‑regularized.

\begin{figure*}[h]
    \centering
    \includegraphics[width=1\linewidth]{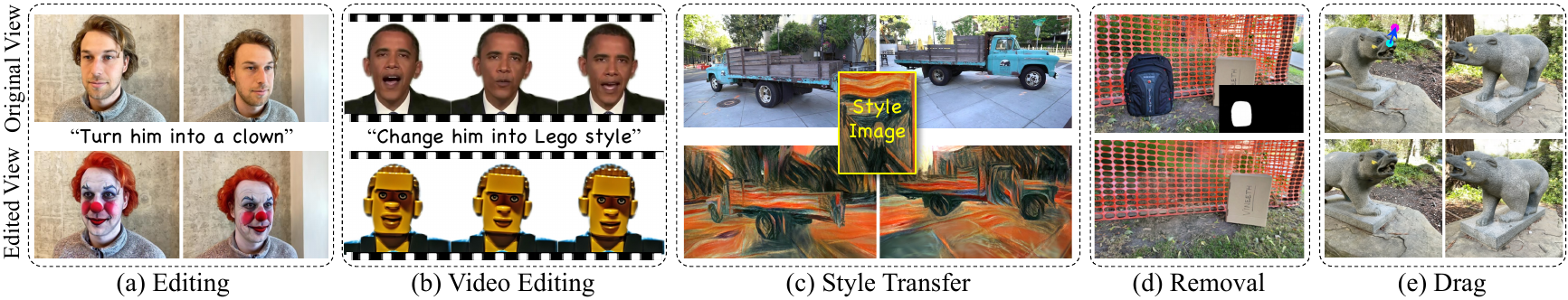}
    \vspace{-6mm}
    \caption{
    Examples of different editing operations in 3DGS, with examples adapted from GaussianEditor~\cite{GaussianEditor-HGS}, PortraitGen~\cite{gao2024portrait}, SGSST~\cite{galerne2024sgsst}, GScream~\cite{GScream}, and DYG~\cite{qu2025drag}.    
    }
    \label{fig:edit_task}
\vspace{-12pt}
\end{figure*}

\noindent $\bullet$ \textbf{Joint Optimization of Multiple Cues.}
To enhance the precision of 2D mask lifting, recent methods jointly optimize semantic, appearance, and geometric cues end-to-end.
COB-GS~\cite{COB-GS} proposes a joint optimization framework that simultaneously learns semantics and texture, leading to sharper object boundaries while preserving visual fidelity.
PanoGS~\cite{PanoGS} formulates a language-guided graph cut mechanism that aggregates 3D Gaussian primitives into super-primitives by leveraging both reconstructed geometry and linguistic signals. It further applies graph-based clustering with SAM-guided edge affinity to produce 3D-consistent panoptic segmentations.
InstanceGaussian~\cite{InstanceGaussian} mitigates the mismatch between appearance and semantics by progressively co-training both modalities within the Scaffold-GS~\cite{Scaffold-gs} framework, achieving better alignment and instance-level coherence.

\noindent $\bullet$ \textbf{Training-Free Mask Lifting.}
To avoid the computational cost and data requirements of model training, several methods explore training-free strategies that directly associate 2D segmentation masks with 3D Gaussian primitives for efficient 3D segmentation.
SAGD~\cite{SAGD} adopts a simple yet effective projection-based strategy: the centers of 3D Gaussians are projected onto 2D masks, and those falling within foreground regions are classified as foreground Gaussians.
FlashSplat~\cite{FlashSplat} formulates the mask lifting process as a one-step linear programming (LP) optimization problem, directly transforming 2D mask supervision into 3D Gaussian masks without iterative training.
GaussianCut~\cite{GaussianCut} represents the scene as a graph over Gaussians and applies graph-cut optimization to partition them into foreground and background. It leverages coarse segmentation from 2D image/video models and refines the results through learned edge affinities in the constructed graph.
THGS~\cite{THGS} constructs a hierarchical superpoint graph from 3D Gaussian primitives and reprojects 2D semantic features onto this structure, producing a view-consistent semantic field without the need for supervised training.
ReLaGS~\cite{ReLaGS} builds upon THGS by constructing an open-vocabulary 3D semantic scene graph for efficient 3D perception and relational reasoning without scene-specific training.
iSegMan\cite{iSegMan} introduces a visibility-guided voting scheme that links 2D segmentations from SAM with 3D Gaussians, treating the association as a voting process weighted by Gaussian visibility.
LBG~\cite{LBG} assigns semantics to 3D Gaussians via a 2D-to-3D lifting and incrementally merges them based on semantic and geometric overlap, enabling fast training-free 3D instance segmentation.
LUDVIG~\cite{LUDVIG} introduces ``inverse rendering'' aggregation, where 2D features or masks from multiple views are uplifted to the 3D by performing a weighted average guided by the 3DGS rendering weights. Then, a graph diffusion mechanism is designed for feature refinement.

\subsubsection{Feed-Forward-based Methods}\label{sec:3.1.3}
To overcome the inefficiency of per-scene optimization, especially under dense calibrated images conditions, several methods~\cite{SLGaussian,GSemSplat,DrSplat,SemanticSplat,LangScene-X,SpatialSplat} adopt feed-forward architectures to enable fast and generalizable 3D semantic field construction.
LSM~\cite{LSM} is the first feed-forward framework that integrates DUSt3R~\cite{wang2024dust3r}, point transformer~\cite{pointtransformer}, and LSeg~\cite{Lseg} for language-driven scene understanding. 
Following LSM's paradigm, SIU3R~\cite{siu3r} is proposed to learn scene understanding and 3D reconstruction simultaneously without feature alignment.
SLGaussian~\cite{SLGaussian} builds a feed-forward pipeline that infers 3D semantic Gaussians directly from sparse viewpoints based on MVSplat~\cite{chen2024mvsplat}. 
Dr. Splat~\cite{DrSplat} departs from traditional rendering-based supervision by directly associating CLIP embeddings with 3D Gaussians. It introduces a language feature registration mechanism that assigns CLIP features to dominant Gaussians intersected by pixel rays. Additionally, it employs Product Quantization (PQ) learned from large-scale image datasets to compress features.
LangScene-X~\cite{LangScene-X} first uses a TriMap video diffusion network based on CogVideoX~\cite{CogVideoX} to densify sparse inputs, synthesizing appearance, geometry, and semantics. It then employs a Language‑Quantized Compressor (LQC) to encode language embeddings on large-scale datasets.
SceneSplat~\cite{SceneSplat} introduces a self-supervised framework that facilitates rich 3D feature learning from unlabeled scenes. To support this effort, SceneSplat-7K is introduced as the first large-scale dataset specifically designed for 3DGS in indoor environments.

\subsubsection{Related Task: 3DGS Object Detection}\label{sec:object_detection}
3D object detection focuses on identifying and localizing objects in 3D space.
Similar to segmentation, detection methods can be categorized into prompt-based and generic approaches. 
Prompt-based detection, often referred to as open-vocabulary detection or visual grounding, takes textual inputs (\eg, category names or language expressions) and outputs object locations. Most 3DGS-based segmentation methods also support this type of detection by generating the maximum response point, and are thus not detailed here.
Specifically, SpatialReasoner~\cite{SpatialReasoner} combines LLM-driven spatial reasoning with a hierarchical feature field enhanced by visual properties. It distills CLIP and SAM features and leverages fine-tuned LLMs to infer spatial instructions, enabling precise localization of target instances based on relational language cues.
ReasonGrounder~\cite{liu2025reasongrounder} leverages large vision-language models (LVLM) to interpret both explicit and implicit instructions. It localizes objects, even when partially or fully occluded, via hierarchical 3D feature fields that enable scale-aware Gaussian grouping.
GVR~\cite{liao2025zero} is a zero-shot 3D visual grounding framework that reformulates 3DVG as a 2D retrieval task. Relying on existing 2D visual foundation models and a built knowledge book, it removes the need for 3D annotations and per-scene training.
For generic object detection, representative methods include:
Gaussian-Det~\cite{Gaussian-Det}, which models objects continuously using input Gaussians as feature descriptors across partial surfaces; 3DGS-DET~\cite{3dgs-det}, which incorporates boundary guidance and box-focused sampling to refine Gaussian distributions and suppress background noise; and MATT-GS~\cite{MATT-GS}, which employs masked attention 3D Gaussian Splatting to enhance localization.

\subsection{3DGS Editing}\label{sec:3.2}
Inspired by 2D editing methods~\cite{brooks2023instructpix2pix,yang2025psp}, editing works based on 3D Gaussian Splatting (3DGS) have advanced~\cite{SC-GS,3D-Gaussian-Editing-with-A-Single-Image,xie2024physgaussian}. These methods leverage the rich scene representation of 3DGS and often incorporate 2D diffusion models to guide precise 3D modifications. This section reviews and categorizes editing techniques and tasks, with an illustrative overview provided in Fig.~\ref{fig:edit_task}.

\subsubsection{Editing Based on Text Prompt}\label{sec:3.2.1}
Text-Driven 3D Editing \cite{EditSplat,GSEdit,huang2025dacapo} represents an emerging and rapidly advancing technology that allows users to enable convenient and efficient editing or generation of 3D scenes through simple natural language instructions, significantly reducing the technical barriers to 3D modeling and content creation. 

\noindent $\bullet$ \textbf{Localizing the Editing Object}. Most current research first requires text to locate the objects that need to be edited. GaussianEditor~\cite{GaussianEditor-HGS,GaussianEditor-ROI}, as pioneering work in the field of 3D editing based on 3D Gaussian Splatting technology, proposes innovative editing methodologies. The former significantly enhances editing precision by integrating textual descriptions throughout the training process to track editing targets. The latter achieves refined control over the editing process by extracting regions of interest (ROIs) corresponding to textual instructions and mapping these regions to 3D Gaussians.
These methods leverage the InstructPix2Pix~\cite{brooks2023instructpix2pix} model during the editing process, but due to its inherent limitations, issues like inaccurate localization and limited editing control still persist. To address these, GSEditPro~\cite{GSEditPro} introduces an attention-based progressive localization module, leveraging cross-attention layers from the T2I model to classify Gaussians and precisely localize editing regions. It also incorporates DreamBooth~\cite{ruiz2023dreambooth} for enhanced diversity and flexibility. Xiao \etal~\cite{xiao2025localized} propose a lighting-aware 3D scene editing pipeline, utilizing an Anchor View Proposal (AVP) algorithm to identify target regions and a coarse-to-fine optimization process with Depth-guided Inpainting Score Distillation Sampling (DI-SDS) for texture and lighting consistency. RoMaP~\cite{kim2025robust} employs 3D-Geometry Aware Label Prediction to identify editing targets and combines SDS loss with additional regularization to produce high-quality local edits while preserving contextual consistency.

\noindent $\bullet$ \textbf{Efforts toward Multi-View Consistency}. Editing 3D objects while maintaining multi-view consistency is a core challenge in the field of 3D editing. 
GaussCtrl~\cite{GaussCtrl} leverages depth-conditioned editing by utilizing naturally consistent depth maps to ensure geometric consistency across multi-view images, along with attention-based latent code alignment to achieve a unified appearance across views. Luo~\etal~\cite{TrAME} propose a progressive 3D editing strategy using a Trajectory-Anchored Scheme (TAS) and a dual-branch editing mechanism to ensure multi-view consistency, addressing challenges in error accumulation during text-to-image processes. Gomel~\etal~\cite{gomel2024diffusion} introduce a geometry-guided warping mechanism that leverages the scene’s depth and structural information to accurately map edits across views, ensuring multi-view consistency. DGE~\cite{DGE} employs spatiotemporal attention to jointly edit the selected key views and inject their features into other views, using correspondences derived from epipolar-constrained visual features, thereby establishing multi-view consistency. SplatFlow~\cite{go2024splatflow} leverages a multi-view rectified flow (RF) model to simultaneously generate multi-view images, depth information, and camera poses based on text prompts. By incorporating training-free inversion and inpainting techniques, it achieves seamless editing of 3DGS. InterGSEdit~\cite{wen2025intergsedit} introduces a 3D Geometry-Consistent Attention Prior and an Adaptive Cross-Dimensional Attention Fusion Network to ensure multi-view consistency and enable fine-grained detail recovery. Pro3D-Editor~\cite{zheng2025pro3d} introduces Primary-view Sampler and Key-view Render to prevent inconsistency conflicts during multi-view editing. DFFSplat~\cite{kohdiffusion} integrates a 3D-consistent diffusion feature field into the editing pipeline and uses a dual-encoder architecture to disentangle view-independent structure from view-dependent appearance, enabling consistent multi-view editing.

\noindent $\bullet$ \textbf{Efforts toward Efficiency and Speed}. Additionally, to enhance the efficiency and speed of 3D editing, ProGDF~\cite{ProGDF} comprises Progressive Gaussian Sampling (PGS) and Gaussian Difference Field (GDF). PGS applies progressive constraints to generate diverse intermediate results during the editing process. GDF employs a lightweight neural network, leveraging these intermediate results to model the editing process, allowing for real-time, controllable, and flexible editing within a single training session. The optimization process of DreamCatalyst~\cite{kim2024dreamcatalyst} approximates the reverse diffusion process, aligning with diffusion sampling dynamics and thereby reducing training time. 3DSceneEditor~\cite{3DSceneEditor} utilizes a streamlined 3D pipeline, allowing direct manipulation of Gaussians. By leveraging input prompts, semantic labeling, and CLIP’s zero-shot capabilities, it enables efficient, high-quality editing. Chen \etal~\cite{DGE} propose Direct Gaussian Editor, which adapts 2D image editing models into a multi-view consistent version and integrates the 3D geometric representation of the underlying scene. This method enables direct optimization of the 3D representation, eliminating the need for iterative editing. 3DitScene~\cite{3DitScene} incorporates generative priors and optimization techniques to refine the 3D Gaussian model and utilizes language features extracted by CLIP for object disentanglement. 

\subsubsection{Editing Based on Image Prompt}\label{sec:3.2.2}
Editing results that rely solely on text prompts often fail to fully meet user expectations, as textual descriptions may not sufficiently convey user preferences regarding details, styles, or specific appearances. To address this, some methods \cite{GScream}, \cite{waczynska2024d}, \cite{Texture-GS, he2025ctrl} have begun incorporating image prompts as a supplement. Image prompts offer more intuitive and customized visual information, including color, texture, shape, and lighting details, enabling more precise visual guidance in the 3D editing process. 

\noindent $\bullet$ \textbf{Parameter-Efficient Fine-Tuning}\label{3_2_2_1}.  To achieve personalized customization, it is necessary to learn from reference images and transfer them to the target object. TIP-Editor~\cite{TIP-Editor} introduces a progressive 2D personalization strategy that incorporates localized loss to ensure edits remain confined to user-defined regions. It employs LoRA~\cite{LoRA} technology to fine-tune the text-to-image (T2I) model, binding the reference image to specific tokens for content personalization. Additionally, it utilizes explicit and flexible 3D Gaussian splatting as the 3D representation, enabling localized editing without altering the background. With a similar idea of \cite{GS-VTON}, GS-VTON introduces a reference-driven image editing approach that integrates personalized information into a pre-trained 2D VTON model using LoRA fine-tuning. This method enables multi-view image editing while ensuring consistency across all views. In contrast, Cha et al.~\cite{cha2024perse} propose PERSE, which learns a disentangled latent space from synthetic data to enable intuitive 3D facial editing, allowing attribute transfer from a reference image while preserving identity.
Texture-GS~\cite{Texture-GS} employs a UV mapping MLP, a local Taylor expansion of the MLP, and a learnable texture to decouple appearance from geometry by representing it as a 2D texture on a 3D surface.

\noindent $\bullet$ \textbf{Improving Multi-View Consistency with Diffusion Models.
} Given their robust theoretical foundation and exceptional performance, diffusion models have naturally been widely adopted in the field of 3D editing. VcEdit~\cite{vcEdit} introduces the Cross-attention Consistency Module and the Editing Consistency Module, integrating 3DGS into the image editing process. This ensures that the edited guidance images maintain consistency across multiple views. Similarly, TIGER~\cite{TIGER} introduces Coherent Score Distillation, which combines a 2D image editing diffusion model with a multi-view diffusion model to enhance score distillation, enabling multi-view consistent editing with finer details.

\noindent $\bullet$ \textbf{Multi-Stage Progressive Refinement}. The multi-stage approach, through phased processing, enables the model to preserve important details of the original image while enhancing the editing effects in the target regions, thereby better adapting to complex editing tasks. Point’n Move~\cite{Point'n-Move} additionally introduces 2D hint points as conditions and designs a two-stage self-prompting segmentation algorithm for mask refinement and merging, enabling real-time editing without requiring training for each edit. Szymkowiak~\etal~\cite{szymkowiak2024neural} design a three-stage method: it begins with a set of input images and camera poses, utilizing a neural signed distance field (SDF) to reconstruct the scene surface, which guides the training of Gaussian splatting components to ensure alignment with the scene geometry. Finally, visual and geometric information is encoded into a lightweight triangle soup proxy. Edits are propagated through this intermediate structure to the mesh extracted from neural surface, thereby updating restored appearance. GaussianVTON~\cite{GaussianVTON} pioneers a 3D Virtual Try-On pipeline by integrating Gaussian Splatting with 2D VTON, introducing a three-stage refinement strategy to enhance consistency.

\subsubsection{Style Transfer}\label{sec:3.2.3}
Unlike object manipulation tasks, style transfer requires preserving the structural integrity of the object while transferring the stylistic features from a reference image to the target image. This process typically involves the separation and recombination of low-level features (such as color and texture) and high-level features (such as semantic content) of the image. 

\noindent $\bullet$ \textbf{Efforts toward Objective Function}. SGSST~\cite{galerne2024sgsst} designs a Simultaneously Optimized Scales loss, enabling style transfer across all scales for consistent multi-scale stylization.
ReGS~\cite{Regs} improves stylization consistency via a Stylized Pseudo View Supervision loss for uniform appearance across views, and a Template Correspondence Matching loss to propagate style to occluded regions. WaSt-3D~\cite{kotovenko2024wast} performs style transfer directly on 3D Gaussians, using an entropy-regularized Wasserstein-2 distance to smoothly map the style scene distribution to the content scene via gradient flow, and decomposing stylization into smaller subproblems for efficiency. Multi-StyleGS~\cite{lin2025multi} introduces a style loss with bipartite matching between multiple style image regions and GS points to enable local style transfer. CLIPGaussian~\cite{howil2025clipgaussian} jointly leverages CLIP and VGG features to enable style transfer while preserving the original content. 
GT$^2$-GS~\cite{GT2-GS} models texture-level style via a geometry-aware transfer loss with cross-view priors for view-consistent stylization. ABC-GS~\cite{ABC-GS} aligns rendering and style features in feature space to better capture global style consistency.

\noindent $\bullet$ \textbf{Efforts toward Integrating Diffusion Model}. 
InstantStyleGaussian~\cite{InstantStyleGaussian} accelerates style editing by using a diffusion model to generate target style images, incorporating them into the training set, and iteratively optimizing the Gaussian splatting scene. ArtNVG~\cite{ArtNVG} enhances control over content and style via the CSGO model and Tile ControlNet, and introduces an Attention-based Neighboring-View Alignment mechanism to maintain consistent colors and textures across neighboring views. Morpheus~\cite{wynn2025morpheus} employs an RGBD diffusion model with depth-guided cross-attention, feature injection, and a Warp ControlNet conditioned on composite frames to guide 3D stylization. FantasyStyle~\cite{yang2025fantasy} proposes Controllable Stylized Distillation, achieving 3D style transfer solely through diffusion model distillation.
DiffStyle3D~\cite{yang2026diffstyle3d} performs style transfer by directly computing the distance to the style image within the self-attention feature space of the diffusion model.

\noindent $\bullet$ \textbf{Efforts toward VGG Feature Optimization}. 
Saroha~\etal~\cite{saroha2024gaussiansplattingstyle} employ a pre-trained VGG model with AdaIN to transfer styles to specific views, training a 3D color module to predict new colors for each Gaussian. Similarly, StyleSplat~\cite{StyleSplat} refines style transfer by applying a nearest neighbor feature matching loss between VGG features of the rendered image and the reference style image, and uses a 2D mask for selective stylization of specified objects. In contrast, StyleGaussian~\cite{StyleGaussian} embeds 2D VGG scene features into transformed 3D Gaussian features, combines them with reference image features, and decodes them into stylized RGB images. SemanticSplatStylization~\cite{SemanticSplatStylization} integrates semantic understanding and uses VGG features to compute style loss, ensuring precise, context-aware stylization.

\noindent $\bullet$ \textbf{Feed-Forward-based Model}. Styl3R~\cite{wang2025styl3r} introduces a branched architecture that separates structural modeling from appearance shading, effectively preventing style transfer from distorting the underlying 3D scene structure, while employing an identity loss to facilitate training of the stylized model. $\text{A}^{3}$GS~\cite{fang2025a3gs}, a novel feed-forward framework, introduces an AdaIN-based stylizer that effectively injects the style features of a target image into the latent representations of a 3DGS scene.
Stylos~\cite{liu2025stylos} uses a Transformer with two paths: self-attention for accurate geometry, and cross-attention to apply consistent style across views.

\subsubsection{Other Editing Tasks}\label{sec:3.2.4}
$\bullet$ \textbf{Object Removal}. Gaussian Grouping~\cite{GS-Grouping} uses 2D mask predictions from the SAM~\cite{SAM} to assign a compact identity encoding to each Gaussian, enabling direct manipulation of specified ones. In contrast, GScream~\cite{GScream} introduces multi-view monocular depth estimation as an additional constraint to optimize the placement of Gaussian primitives, improving the geometric alignment between the removed area and the surrounding region. Qiu~\etal~\cite{qiu2024feature} propose the Feature Splatting method, which distills vision-language features into a 3D Gaussian representation to enable semi-automatic scene decomposition via text queries, thereby facilitating scene editing operations such as object removal.

\noindent $\bullet$ \textbf{Drag}. MVDrag3D~\cite{chen2024mvdrag3d} employs a multi-view diffusion model as a robust generative prior to achieve consistent drag editing across multiple rendered views. DYG~\cite{qu2025drag} uses 3D masks and control point pairs to define the target editing area and drag direction, leveraging the advantages of implicit tri-plane representation to establish a geometric framework for the editing results.
3DGS-Drag~\cite{dong20263dgs} enables intuitive 3D drag editing by combining deformation guidance for consistent geometric changes and diffusion guidance for content refinement.

\noindent $\bullet$ \textbf{Video Editing}. 
3DEgo~\cite{khalid20243dego} introduces a noise blending module to apply the diffusion model for video frame editing and leverages 3D Gaussian splatting to generate 3D scenes from multi-view consistent edited frames, reducing the multi-stage editing workflow to a single-stage process. PortraitGen~\cite{gao2024portrait} lifts 2D portrait video editing into 3D by integrating 3D human priors, ensuring both 3D and temporal consistency in the edited video.

\noindent $\bullet$ \textbf{Inpainting}. InFusion~\cite{InFusion} uses an image-conditioned depth completion model to guide point initialization and directly recovers depth maps from images. RefFusion~\cite{RefFusion} fine-tunes an image inpainting diffusion model, effectively aligning the prior distribution with the 3D target scene. This reduces score distillation variance and yields clearer details. PAInpainter~\cite{cheng2025perspective} integrates inpainting view sampling, cross-view content propagation, and consistency verification to enable 3D Gaussian inpainting.
\subsection{3DGS Generation}
\label{sec:3.3}
\begin{figure*}[h]
    \centering
    \includegraphics[width=1\linewidth]{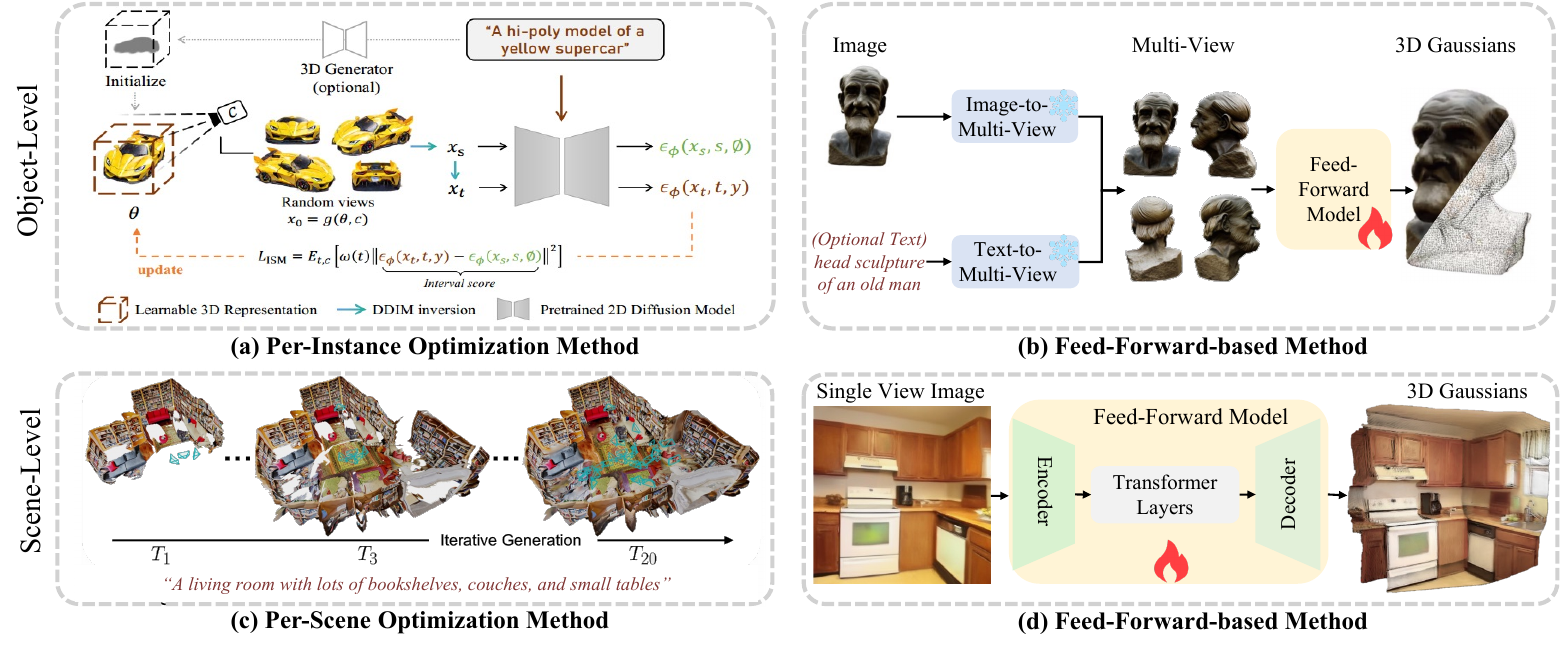}
    \vspace{-6mm}
    \caption{
    A comparison of four 3DGS generation pipelines. The examples are adapted from LucidDreamer~\cite{liang2024luciddreamer}, Text2Room~\cite{hollein2023text2room}, LGM~\cite{LGM}, and CATSplat~\cite{roh2024catsplat}.}
    \label{fig:generation_task}
\vspace{-12pt}
\end{figure*}

While NeRF-based methods produce high-quality 3D content, they are slow and computationally expensive. Recent advances in 3D Gaussian Splatting (3DGS) enable faster, more efficient 3D generation from text or images, categorized by the type of output.
\figurename~\ref{fig:generation_task} provides representative generation pipelines.

\subsubsection{Object-level Generation}\label{sec:3.3.1}
\noindent \textit{\textbf{A. Per-Instance Optimization Methods}}:
Optimization-based methods extract diffusion priors from powerful 2D foundation diffusion models to guide and optimize 3D representations. DreamFusion~\cite{poole2022dreamfusion} first introduces the Score Distillation Sampling (SDS) loss, which optimizes NeRF~\cite{NeRF} using images generated from text prompts. Building on this, many methods~\cite{ wang2024textto3dgaussiansplattingphysicsgrounded,yang2024hash3d,zhou2025layoutdreamer,qiu2024richdreamer} have extensively explored optimization-based approaches. 

\noindent $\bullet$ \textbf{Standard SDS-based Methods}. These methods directly use Score Distillation Sampling (SDS) to optimize 3D scenes.
\textit{(a) Text-to-3D.} Tang\etal~\cite{Dreamgaussian} extend 2D diffusion models to 3D by optimizing 3D Gaussians with SDS, introducing Mesh Extraction and Texture Refinement to mitigate SDS-induced blurriness. GSGEN~\cite{chen2024textto3dusinggaussiansplatting} and GaussianDreamer~\cite{yi2024gaussiandreamer} incorporate 3D point cloud diffusion priors for geometry and appearance, while GaussianDreamerPro~\cite{GaussianDreamerPro} adopts a geometry-guided framework to control Gaussian growth and enhance details. CompGS~\cite{ge2024compgs} decomposes scenes into entities for SDS optimization at entity and composition levels, dynamically adjusting spatial parameters for fine-grained details. CG3D~\cite{CG3D} introduces a compositional framework for text-conditioned scene generation, producing detailed, multi-object, and plausible results. Hyper-3DG~\cite{Hyper-3DG} refines Gaussians via a Geometry and Texture Hypergraph Refiner, and HCoG~\cite{qin2025apply} hierarchically generates occlusion-aware assets.
\textit{(b) Image-to-3D.} ScalingGaussian~\cite{ScalingGaussian} first generates point clouds with a 3D diffusion model, then refines Gaussians using a 2D diffusion model with SDS. Physics3D~\cite{liu2024physics3d} simulates Gaussians via the Material Point Method (MPM) to estimate physical attributes and orientations, rendering them into video frames for SDS-based optimization with a pre-trained video diffusion model.

\noindent $\bullet$ \textbf{Efforts toward Improving SDS}.
Although SDS provides guidance for 3D generation, it often suffers from low fidelity and over-smoothing. Recent works address these issues by enhancing geometric accuracy, detail preservation, and stability.
\textit{(a) Text-to-3D.} StableDreamer~\cite{guo2023stabledreamer} formulates the SDS prior and L2 reconstruction loss as equivalent, reducing multi-face artifacts and improving geometry. Li~\etal~\cite{li2024connecting} propose Guided Consistency Sampling integrated with 3DGS to enhance detail and fidelity. DreamMapping~\cite{cai2024dreammapping} introduces Variational Distribution Mapping, accelerating distribution modeling by treating rendered images as degraded diffusion outputs. HumanGaussian~\cite{HumanGaussian} designs a structure-aware SDS for joint optimization of human appearance and geometry. LucidDreamer~\cite{liang2024luciddreamer} mitigates over-smoothing via Interval Score Matching (ISM) with deterministic diffusion trajectories, while TSM~\cite{miao2024dreamer} extends ISM by generating two reverse DDIM~\cite{song2020denoising} paths to reduce cumulative errors and pseudo-ground truth inconsistencies. GaussianMotion~\cite{shim2025gaussianmotion} employs adaptive score distillation to balance realism and smoothness.
\textit{(b) Image-to-3D.} Basak~\etal~\cite{basak2024enhancingsingleimage3d} introduce a frequency-based distillation loss, extracting low-frequency geometry from 3D diffusion priors and refining high-frequency texture with 2D diffusion. GECO~\cite{GECO} adopts a two-stage pipeline combining Variational Score Distillation~\cite{wang2023prolificdreamer} and multi-view consistency refinement, enabling sub-second high-quality generation. DreamPhysics~\cite{DreamPhysics} leverages video diffusion priors to learn material properties and applies motion distillation sampling to emphasize dynamic cues. TraCe~\cite{li2025walking} reformulates the Schrödinger Bridge into an explicit diffusion bridge linking the current rendering to its text-conditioned, denoised target, while training a LoRA-adapted model on the trajectory’s score dynamics for robust 3D optimization.

\noindent $\bullet$ \textbf{Multi-View Guidance}.
The imprecise guidance of SDS often leads to the Janus problem, characterized by multi-face ambiguity. MVGaussian~\cite{MVGaussian}, GradeADreamer~\cite{ukarapol2024gradeadreamer}, and GALA3D~\cite{GALA3D} mitigate this issue by combining SDS with MVDream~\cite{shi2023mvdream} as a multi-view diffusion prior. GALA3D further incorporates layout-guided Gaussian representations and instance-level scene optimization to achieve high-fidelity object generation and realistic scene interactions. Li~\etal~\cite{li2024controllabletextto3dgenerationsurfacealigned} also adopt MVDream with hybrid Gaussian–mesh representations, introducing MVControl and SuGaR for controllable text-to-3D generation.

\noindent \textit{\textbf{B. Feed-Forward-based Methods}}:
Although these optimization-based methods have made significant progress in 3D content generation, they still face challenges such as lengthy optimization times. To address these issues, researchers have begun to shift toward feed-forward methods~\cite{han2024vfusion3d},~\cite{GVGEN}, which involve training on large-scale datasets to directly generate 3D assets. 

\noindent$\bullet$~\textbf{Latent-space Optimization.} 
These methods learn a latent space from conditional inputs such as RGB images, depth maps, or point clouds via an encoder, which is then decoded into 3D Gaussians.
\textit{(a) Text-to-3D.} Wizadwongsa~\etal~\cite{wizadwongsa2024taming} employ a feed-forward reconstruction encoder with pre-trained models to reduce training cost, refining unstructured latents via post-processing. They introduce a 2D perceptual rendering loss and a multi-stream transformer correction flow for efficient, high-quality text-conditioned generation. GaussianAnything~\cite{GaussianAnything} proposes a scalable framework with a point cloud-structured latent space, supporting multi-modal inputs, geometry–texture disentanglement, and 3D-aware editing. Atlas-Gaussians~\cite{Atlas-Gaussians} use a patch-based encoder–decoder with UV sampling and a transformer decoder for efficient, high-fidelity Gaussian shape generation. Turbo3D~\cite{hu2024turbo3d} integrates a four-step, four-view latent diffusion generator with a feed-forward Gaussian reconstructor, employing a dual-teacher scheme for view consistency and photorealism while avoiding pixel-space decoding.
\textit{(b) Image-to-3D.} AGG~\cite{AGG} leverages a pre-trained DINOv2~\cite{oquab2023dinov2} encoder and dual transformers to decode Gaussian position and texture fields for joint optimization. HumanSplat~\cite{pan2024humansplat} encodes multi-view inputs via a VAE and applies a latent reconstruction transformer interacting with structured human models. Zou~\etal~\cite{zou2024triplane} design a hybrid triplane Gaussian latent representation from images and point clouds, decoded into 3D objects with transformer-based point and triplane decoders.
SAM 3D~\cite{chen2025sam} generates 3D structures in the latent space and then decodes them back into explicit 3D representations.

\noindent$\bullet$~\textbf{Multi-View-based Methods.}
Incorporating multi-view inputs can substantially improve 3D generation quality.
\textit{(a) Text-to-3D.} Flex3D\cite{han2024flex3d} adopts a two-stage framework: first, a fine-tuned multi-view image diffusion model and a video diffusion model generate a candidate view pool, filtered for quality and consistency via a view selection pipeline; second, selected views are processed by a transformer-based Flexible Reconstruction Model (FlexRM) that outputs 3D Gaussians using tri-plane representations for efficient, detailed generation. LGM~\cite{LGM} employs an asymmetric U-Net for high-resolution 3D Gaussian splatting, enabling expressive, scalable reconstruction from multi-view inputs without triplanes or transformers.
\textit{(b) Image-to-3D.} GS-LRM~\cite{GS-LRM} uses a transformer architecture that patchifies posed images, processes concatenated multi-view tokens through transformer blocks, and decodes per-pixel Gaussian parameters for differentiable rendering. GEOGS3D~\cite{GeoGS3D} integrates orthogonal plane decomposition with a diffusion model to synthesize geometry-aware, multi-view consistent novel views, enhancing 3D object reconstruction.

\noindent$\bullet$~\textbf{Network Design.} 
Designing feed-forward networks tailored to 3DGS is an active research direction.
\textit{(a) Text-to-3D.} BrightDreamer \cite{BrightDreamer} proposes a feed-forward framework combining a text-guided shape deformation network with a triplane generator to predict and optimize Gaussian attributes.
\textit{(b) Image-to-3D.} Lu~\etal~\cite{lu2024large} present the Point-to-Gaussian Generator composed of multiple APP blocks, each with a point feature extractor, projection module, and cross-modal attention, followed by multi-linear heads to decode 3D Gaussians. GRM~\cite{GRM} introduces an upsampler using a windowed self-attention to capture non-local cues and enhance high-frequency details. UniGS~\cite{wu2024unigs} is a DETR-like model treating Gaussians as queries, updated via multi-view cross-attention over input images to reduce ghosting artifacts.

\noindent$\bullet$~\textbf{Optimization Diffusion Models.} 
These methods directly produce 3D Gaussians by fine-tuning or training diffusion models.
\textit{(a) Text-to-3D.} GVGEN\cite{GVGEN} adopts a structured Gaussian volume representation and a coarse-to-fine pipeline, using a candidate pool strategy for pruning and densification to reconstruct high-fidelity 3D scenes from complex text.
\textit{(b) Image-to-3D.} NovelGS~\cite{NovelGS} generates 3D Gaussians from sparse-view images with novel-view denoising, achieving state-of-the-art reconstruction with consistent, sharp textures. Ouroboros3D~\cite{wen2024ouroboros3d} jointly trains multi-view image generation and 3D reconstruction in a recursive diffusion process, enabling mutual adaptation for robust inference. Cycle3D~\cite{Cycle3D} cyclically integrates 2D diffusion-based generation and feed-forward 3D reconstruction to improve multi-view consistency and texture quality. DiffusionGS~\cite{cai2024bakinggaussiansplattingdiffusion} outputs Gaussian point clouds at each timestep for view-consistent generation from any direction, introducing a scene–object mixed training strategy to enhance scalability. Hi3D~\cite{Hi3D} employs a video diffusion paradigm with 3D-aware priors and video-to-video refinement for consistent multi-view reconstruction. TRIM~\cite{yin2025trim} employs a lightweight selector model to assess latent Gaussian primitives from multiple noise samples and prunes early trajectories by selecting high-quality candidates, thereby accelerating inference. GSV3D~\cite{tao2025gsv3d} utilizes gaussian splatting for geometric distillation within the SVD~\cite{blattmann2023stable} framework, ensuring structural consistency to boost 3D generation quality.

\subsubsection{Scene-level Generation}\label{sec:3.3.2}\label{sec:3.3.3}

\noindent \textit{\textbf{A. Per-Scene Optimization Methods}}: These methods optimize each scene individually to generate high-quality 3D Gaussians.
\textit{(a) Text-to-3D}. DreamScene~\cite{DreamScene} builds on CSD~\cite{yu2023text}, a variant of SDS, by integrating information across multiple timesteps during sampling, obtaining rich semantic guidance from 2D diffusion models.  FastScene~\cite{ma2024fastscene} proposes a Progressive Novel View Inpainting for generating refined views. DreamScene360~\cite{DreamScene360} uses text-based panorama priors for 3D scene generation with 3DGS to maintain multi-view consistency, and mitigates invisibility in single-view inputs via semantic and geometric regularization.
\textit{(b) Image-to-3D.} Zhong~\etal~\cite{zhong2025taming} introduce scene-grounding guidance in video diffusion models to improve sequence consistency and handle extrapolation and occlusion. WonderWorld~\cite{yu2025wonderworld} proposes Fast Layered Gaussian Surfels (FLAGS) for scene generation, incorporating guided depth diffusion to reduce geometric distortion. Scene4U~\cite{huang2025scene4u} combines LLMs and segmentation models to decompose panoramic images into layers, enabling multi-layered 3D scene generation.

\noindent \textit{\textbf{B. Iterative Generation Methods}}:
These approaches iteratively refine 3D Gaussian representations through repeated rendering and optimization to achieve high-quality scene reconstruction.
\textit{(a) Text-to-3D.} Text2Room~\cite{hollein2023text2room} combines monocular depth estimation with a text-conditioned inpainting model, aligning scene frames with existing geometry for 3D reconstruction. Text2Immersion~\cite{ouyang2023text2immersion} progressively generates Gaussian point clouds using 2D diffusion and depth models, followed by refinement via interpolation. RealmDreamer~\cite{shriram2024realmdreamer} initializes points with a text-to-image generator, lifts them into 3D, computes occlusion volumes, and optimizes across multiple views using an image-conditioned diffusion model. WonderJourney~\cite{yu2024wonderjourney} employs LLMs to generate scene descriptions, builds 3D scenes via text-driven point cloud generation, and validates them with VLMs. HoloDreamer~\cite{zhou2024holodreamer} addresses global inconsistency and incompleteness by generating high-resolution panoramas for holistic initialization, followed by rapid 3DGS-based reconstruction for view-consistent, enclosed scenes.
\textit{(b) Image-to-3D.} ViewCrafter~\cite{yu2024viewcrafter} employs a point-conditioned video diffusion model for high-quality 3D scenes. LucidDreamer~\cite{LucidDreamer} alternates between Dreaming, which generates multi-view consistent images guided by a point cloud, and Alignment, which integrates new 3D points into a unified scene, producing realistic 3D Gaussian splats from text, RGB, or RGBD inputs. VistaDream~\cite{wang2024vistadream} enforces multi-view consistency during reverse diffusion sampling to improve temporal coherence. Kang~\etal~\cite{kang2025multi} propose a transformer-based latent diffusion model with explicit view geometry constraints, incorporating warped feature maps, epipolar-weighted source features, Plücker ray maps, and camera poses.

\noindent \textit{\textbf{C. Feed-Forward-based Methods}}: Similar to object-level generation, feed-forward approaches have also been explored.

\noindent$\bullet$~\textbf{Network Design.\phantomsection}
\textit{(a) Text-to-3D}. TextSplat~\cite{wu2025textsplat} introduces Text-Guided Semantic Fusion Module to integrate multi-source semantic features under sentence-level guidance, enhancing geometry semantic consistency.
\textit{(b) Image-to-3D}. PixelSplat~\cite{pixelSplat} employs a multi-view epipolar transformer to address scale ambiguity, while MVSplat~\cite{chen2024mvsplat} and Splatter360~\cite{chen2025splatter} construct planar or spherical cost volumes for improved geometry estimation. MVSplat360~\cite{chen2024mvsplat360} leverages pre-trained video diffusion for 3D-consistent views. GaussianCity~\cite{xie2025generative} adopts BEV-Point as a compact intermediate representation with a spatial-aware decoder. SelfSplat~\cite{kang2025selfsplat} introduces matching-aware pose estimation and depth refinement to ensure geometric consistency. CATSplat~\cite{roh2024catsplat} fuses vision–language text features with point cloud features, and OmniSplat~\cite{lee2025omnisplat} uses a Yin–Yang grid to reduce distortion.

\noindent$\bullet$~\textbf{Optimization Diffusion Models.} 
\textit{(a) Text-to-3D}. Prometheus\cite{yang2024prometheus} sequentially trains GS-VAE and MV-LDM for direct text-to-3D scene generation. VideoRFSplat~\cite{go2025videorfsplat} couples a pose generation model with a pretrained video generator via communication blocks to produce multi-view images and camera poses.
\textit{(b) Image-to-3D}. Wonderland~\cite{liang2025wonderland} integrates diverse camera trajectories via a dual-branch conditioning mechanism for view-consistent latents. GGS~\cite{schwarz2025generative} combines 3D representations with latent video diffusion and a custom decoder to synthesize scenes from feature fields. VideoScene~\cite{wang2025videoscene} distills video diffusion through a 3D-aware leap flow strategy for one-step generation. 
SceneSplatter~\cite{zhang2025scene} adopts a momentum-based paradigm to balance generative priors and existing scene information during generation.
FlashWorld~\cite{li2025flashworld} introduces dual-mode pretraining on a video diffusion model to enable a multi-view model to operate in both MV-oriented and 3D-oriented modes.

\vspace{-1em}
\revise{\section{Challenges and Outlook}\label{sec:challenge_outlook}}

\revise{\subsection{Open Challenges}\label{sec:challenges}}
\noindent$\bullet$~\textbf{Input Artifacts and Reconstruction Robustness.}
3DGS-driven downstream applications rely critically on the quality of the underlying 3D reconstruction. In practice, however, images acquired under varying conditions, camera pose errors, uneven view coverage, and artifacts such as blur and transient objects often lead to inconsistent geometry and appearance. While such imperfections may be tolerable for reconstruction itself, they are amplified in downstream tasks, causing unstable semantics and view-inconsistent editing or generation. Some recent methods mitigate these issues through depth regularization, blur-aware reconstruction, artifact-aware optimization, and more robust photometric modeling~\cite{chung2023depth,zhao2024bad,peng2024bags}, but they typically target reconstruction quality rather than downstream robustness. Pipelines that explicitly account for input artifacts and other acquisition uncertainties and propagate reconstruction reliability to downstream modules remain largely unexplored, limiting applicability in real-world settings.

\noindent$\bullet$~\textbf{Multi-View Consistency.}
A central challenge across segmentation, editing, and generation is maintaining semantic and structural consistency across views. Many methods optimize objectives primarily in image space, enforcing cross-view coherence only implicitly through rendering supervision. This can yield strong single-view accuracy while still exhibiting semantic drift, temporal flicker, or inconsistent editing across viewpoints. Existing methods partially address this issue using cross-view feature aggregation, 3D lifting, geometry-aware attention, or multi-view alignment constraints~\cite{Unified-Lift,GaussianCut,Click-Gaussian,FMLGS}, yet most focus on specific tasks and do not generalize across the full range of downstream applications. Designing scalable, task-agnostic consistency mechanisms that jointly preserve geometric structure and semantic identity across views remains an open problem.

\noindent$\bullet$~\textbf{Data Scarcity and Supervision Quality.}
Progress in 3DGS downstream applications is strongly constrained by the scarcity of reliable 3D annotations. Since dense 3D labels are expensive to obtain, many pipelines lift pseudo labels from 2D foundation models such as SAM~\cite{SAM} and CLIP~\cite{CLIP}. While effective, such supervision inevitably introduces noise, bias, and incomplete coverage, particularly for occluded regions, thin structures, or long-tail categories. Several methods attempt to improve label quality through cross-view voting~\cite{iSegMan}, contrastive consistency learning~\cite{OpenGaussian,CCL-LGS}, or hierarchical mask refinement~\cite{SA3D-GS}, but these remain largely heuristic and scene-specific. Scalable supervision strategies or large-scale 3D data are still lacking.

\noindent$\bullet$~\textbf{Efficiency.}
Despite their strong performance, many 3DGS downstream pipelines remain computationally expensive because they repeatedly optimize a 3D representation and render it back to 2D for supervision. This tight coupling between 3D representation learning and repeated differentiable rendering makes training and adaptation costly, especially for large scenes, long sequences, and interactive settings. Existing methods have begun to address this bottleneck through feed-forward Gaussian prediction, compact representations, pruning and compression, and pretrained reconstruction priors~\cite{szymanowicz2024splatter,chen2024lara,zou2024triplane,lee2023compact,fan2024lightgaussian,chen2024hac}. However, these methods often trade accuracy or robustness for speed, and their benefits are not yet uniform across segmentation, editing, and generation tasks. The challenge is how to achieve a better balance among efficiency, scalability, and downstream performance.

\noindent$\bullet$~\textbf{Evaluation.}
Existing evaluation protocols largely inherit 2D or NeRF-era metrics such as PSNR, SSIM, and LPIPS, which focus on image-level fidelity and may overlook properties critical to downstream applications, including cross-view semantic consistency, geometric correctness, edit locality, and controllability. Recent efforts have started to incorporate human studies, MLLM-based assessment, and task-specific measures~\cite{jayasumana2024rethinking,wu2023human,liu2023visual}. However, evaluation remains fragmented across datasets and applications. A major open problem is to develop standardized, 3D-aware evaluation frameworks that jointly assess geometry, semantics, multi-view consistency, and practical usability in a unified manner.

\vspace{1em}

\subsection{Future Directions}
\label{sec:future_direction}
\noindent$\bullet$~\textbf{Combining 3D Foundation Models.}
To overcome the robustness and multi-view consistency challenges, emerging 3D foundation models such as VGGT~\cite{VGGT} and FLARE~\cite{FLARE} offer robust representations by pre-training on diverse 3D tasks (e.g., depth estimation, point cloud reconstruction). These models encode rich geometric priors that can stabilize downstream applications under sparse or artifact-heavy inputs. Incorporating them into 3DGS pipelines to improve generalization is a promising direction.

\noindent$\bullet$~\textbf{Toward Generalist Models.}
Instead of designing separate models for segmentation, editing, and generation, a promising direction is to develop generalist 3DGS-based architectures that perform multiple scene-level tasks within a unified framework. Such models can facilitate cross-task representation sharing, reduce training redundancy, and improve generalization.

\noindent$\bullet$~\textbf{Integrating Large Language Models.}
Combining 3DGS with Large Language Models (LLMs) opens new opportunities for semantic understanding and instruction-based manipulation, such as text-driven editing, generation, and open-vocabulary understanding. Research is needed to build effective 3D-text alignment and prompting strategies. Leveraging LLMs to reason over spatial relationships and affordances in 3D scenes also presents promising avenues for embodied intelligence and interactive applications.

\noindent$\bullet$~\textbf{Synthesizing 3D Data.}
To mitigate the scarcity of reliable 3D supervision, future research can leverage the generative capabilities of 3DGS to synthesize large-scale, high-quality 3D datasets from abundant 2D image collections. Such synthetic data could significantly enhance training diversity and volume. Combining synthesized data with domain adaptation techniques could further bridge the gap between synthetic and real-world distributions, improving robustness and generalization.

\noindent$\bullet$~\textbf{Large-Scale Feed-Forward Learning.}
To address the efficiency bottleneck of repeated scene-specific optimization, an increasing number of works are moving toward feed-forward architectures. Future research should explore large-scale training regimes across diverse scenarios and domains, including scaling up data volume, improving data diversity (\eg, egocentric, indoor, outdoor), and designing robust pipelines that generalize in the real world.

\noindent$\bullet$~\textbf{Faithful Evaluation Metrics.}
Current evaluation protocols often inherit metrics from traditional 2D or NeRF-based tasks, which may not align well with 3DGS-specific goals. Developing reliable and interpretable metrics that go beyond 2D projection accuracy to capture inherent 3D geometric and semantic consistency is essential for tracking progress and guiding future research.
\vspace{-1.2mm}
\section{Conclusion}
\label{sec:conclusion}
This survey systematically explores recent advances in utilizing 3D Gaussian Splatting (3DGS) for downstream application tasks, representing a pioneering effort in summarizing this emerging field.
We first introduce essential background knowledge, covering fundamental concepts and foundational models.
Subsequently, we comprehensively review over 200 representative methods across several key tasks, including segmentation, editing, generation, and other related application tasks, providing a structured categorization from a technical perspective.
In the experimental section, we detail the evaluation metrics and compare methods under a fair and consistent protocol.%
Lastly, we highlight current challenges and outline promising future directions to inspire continued research efforts toward enhanced high-level downstream tasks.

\vspace{-1em}
\section*{\Large{Appendices}}\label{sec:appendix}
\vspace{4pt}
\setcounter{table}{0}
\setcounter{figure}{0}
\setcounter{section}{0}
\setcounter{subsection}{0}
\renewcommand{\thetable}{A.\arabic{table}}
\renewcommand{\theequation}{A.\arabic{equation}}

\renewcommand{\thesection}{A.\arabic{section}}

\begin{figure}[t]
    \centering
    \includegraphics[width=1\linewidth]{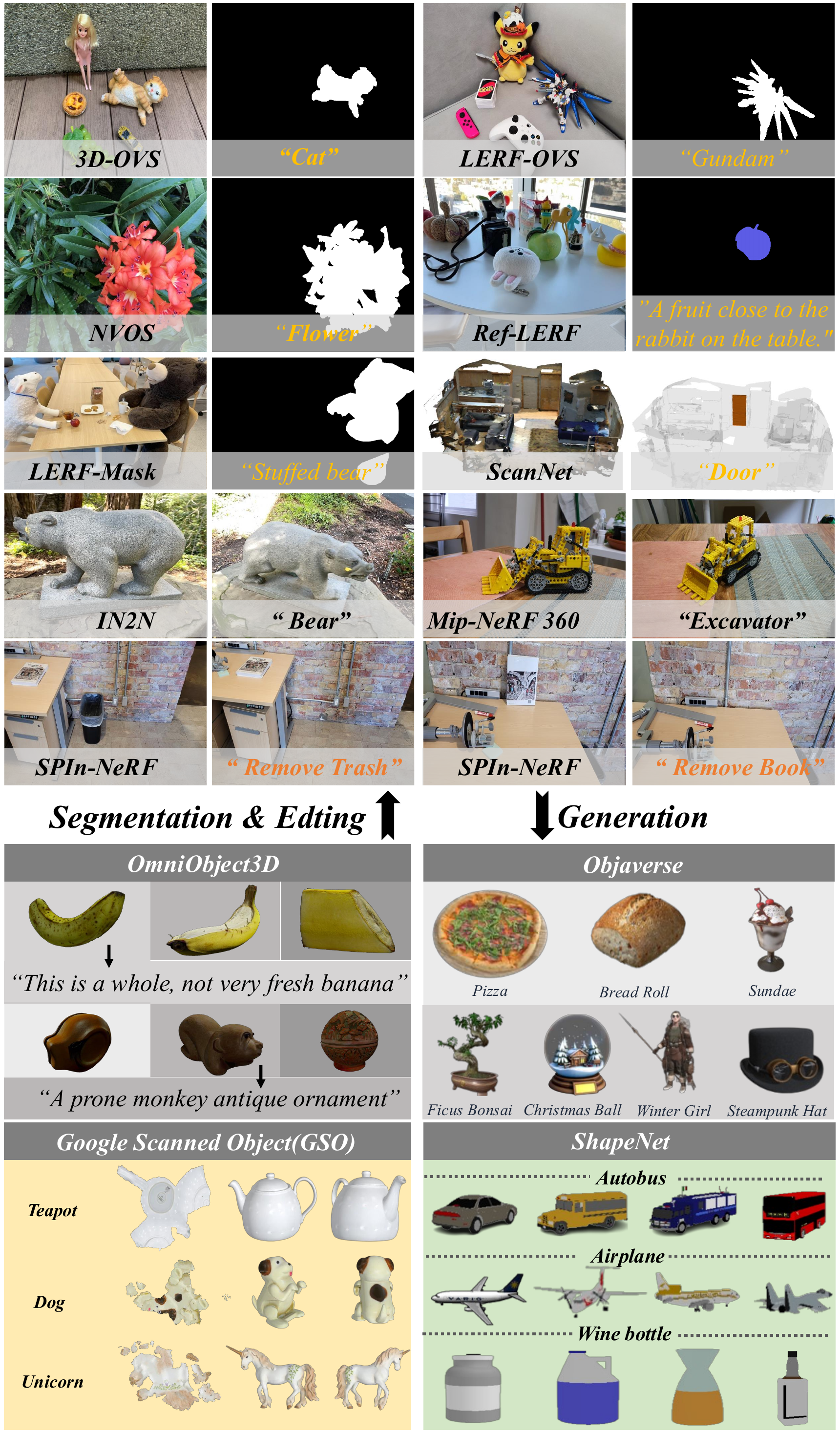}
      \put(-205.5,382){\middlefont~\cite{3D-OVS}}
      \put(-77,382){\middlefont~\cite{LangSplat}}
      \put(-207,330){\middlefont~\cite{NVOS}}
      \put(-77,330){\middlefont~\cite{Ref-LERF}}
      \put(-205.5,288){\middlefont~\cite{GS-Grouping}}
      \put(-79.5,288){\middlefont~\cite{dai2017scannet}}
      \put(-207.5,242.5){\middlefont~\cite{haque2023instructnerf2nerfediting3dscenes}}
      \put(-79,242.5){\middlefont~\cite{barron2022mipnerf360unboundedantialiased}}
      \put(-76,199){\middlefont~\cite{SPIn-NeRF}}
      \put(-206,199){\middlefont~\cite{SPIn-NeRF}}
      \put(-168,170){\middlefont~\cite{OmniSeg3D}}
      \put(-45,170){\middlefont~\cite{deitke2023objaverse}}
      \put(-146,86.5){\middlefont~\cite{GSO}}
      \put(-50,86.5){\middlefont~\cite{chang2015shapenet}}
    \vspace{-3mm}
    \caption{Examples from 13 commonly used datasets for segmentation, editing, and generation.
    }
    \label{fig:dataset}
\vspace{-3.6pt}
\end{figure}
\section{Performance Comparison}
\label{sec:performance}
This section compares different methods across tasks. For each task, we introduce the datasets and evaluation metrics, followed by detailed result tables. Fig.~\ref{fig:dataset} shows examples from 13 commonly used datasets in segmentation, editing, and generation.

\begin{table*}[t]
\centering
\caption{\textbf{Statistics of representative 3DGS segmentation datasets.} See Sec.~\textcolor{red}{4.1} for more detailed descriptions.}

\vspace{-3mm}
\begin{threeparttable}
\fontsize{10pt}{12pt}\selectfont
\resizebox{0.99\textwidth}{!}{
\setlength\tabcolsep{8pt}
\renewcommand\arraystretch{1.0}
\begin{tabular}{|r||c|c|c|l|}
       \hline\thickhline
\rowcolor{mygray}
\textbf{Dataset} & \textbf{Venue} & \textbf{\#Scene} & \textbf{\#Views(Avg.)} & \textbf{Characterization} \\
\hline
\hline

ScanNet~\cite{dai2017scannet} & \mypub{CVPR'17} & 1513 & 1500  & Large-scale RGB-D scans with 3D poses and semantics for advanced scene understanding. \\

\rowcolor{mygray2}
Replica~\cite{straub2019replica} & \mypub{ArXiv'19} & 18 & 175 & High-quality indoor scans with geometry, HDR textures, and rich semantic labels.
 \\
NVOS~\cite{NVOS} & \mypub{CVPR'21} & 8 & 36 & Built on LLFF with undistorted images, annotated with masks and scribbles for segmentation tasks.
\\
\rowcolor{mygray2}%
Mip-NeRF 360\cite{barron2022mipnerf360unboundedantialiased} & \mypub{CVPR'22} & 9 & 215 & Focusing on capturing complex lighting, geometry, and texture details. \\
SPIn-NeRF~\cite{SPIn-NeRF} &\mypub{CVPR'23} & 10 & 100  & Providing challenging real-world scenes with views both with and without a target object.
\\

\rowcolor{mygray2}

3D-OVS\cite{3D-OVS} & \mypub{NeurIPS'23} & 10 & 30 & Including high-quality 3D objects spanning diverse categories with language-aligned semantic labels. \\

LERF-OVS~\cite{LangSplat} & \mypub{CVPR'24} & 4 & 200 & An extended version of LERF dataset with ground truth mask annotations for open-vocabulary segmentation.\\
\rowcolor{mygray2}
LERF-Mask~\cite{GS-Grouping} & \mypub{ECCV'24} & 3 & 200 & Containing semantic annotations of three scenes from LERF dataset~\cite{LERF} with a total of 23 prompts.\\
Ref-LERF~\cite{Ref-LERF} & \mypub{ICML'25} & 4 & 200 & Focusing on spatial relationships, annotated with natural language expressions for referring 3DGS segmentation. \\ 
\rowcolor{mygray2}
SceneSplat-7K~\cite{SceneSplat} & \mypub{ICCV'25} & 7k & - & The first large-scale, high-quality 3DGS dataset for indoor environments boosting scene understanding research. \\ 
SceneSplat-49K~\cite{ma2025scenesplat++}& \mypub{ArXiv'25} & 49k & - &  Containing diverse indoor and outdoor scenes, featuring complex, high-quality full scenes from multiple sources. \\ 
\hline
\end{tabular}
}
\end{threeparttable}
\label{tab:3dgs_segmentation_datasets}
\end{table*}

\vspace{-1.2mm}
\subsection{Performance Benchmarking: 3DGS Segmentation}\label{sec:exp_seg}

\noindent$\bullet$~\textbf{Datasets.}
Table~\ref{tab:3dgs_segmentation_datasets} summarizes the key features of the datasets for 3DGS segmentation.

\noindent\textbf{ScanNet}~\cite{dai2017scannet} contains 1,513 RGB-D sequences from 707 indoor scenes, offering 2.5M frames with 3D reconstructions, camera poses, and dense semantic/instance annotations. It includes ~36K labeled objects across ~20 categories and is a standard benchmark for indoor scene understanding.

\noindent\textbf{Replica}~\cite{straub2019replica} is a high-quality indoor 3D dataset with dense geometry, HDR textures, accurate material properties (\eg, glass, mirrors), planar segmentation, and semantic/instance annotations, supporting realistic simulation and scene understanding.

\noindent\textbf{NVOS}~\cite{NVOS} is a segmentation-focused extension of LLFF with undistorted images, mask and scribble annotations, designed for language-guided 2D-to-3D segmentation.

\noindent\textbf{Mip-NeRF}~\cite{barron2022mipnerf360unboundedantialiased} provides unbounded real-world scenes with precise poses for 360° novel view synthesis.

\begin{table}[t]
\centering
\caption{\textbf{Quantitative 3D instance segmentation on Replica\cite{straub2019replica} and LERF-Mask\cite{GS-Grouping}}. Following~\cite{Unified-Lift}, we report mIoU and F-score on Replica, and mIoU and mBIoU on LERF-Mask.} 

\vspace{-3mm}
\footnotesize    
    \setlength{\tabcolsep}{1mm}
\renewcommand\arraystretch{1.0}
\resizebox{0.498\textwidth}{!}{
\begin{tabular}{|r||c|c|cc|cc|}
       \hline\thickhline
\rowcolor{mygray}
 & & &  \multicolumn{2}{c|}{\textbf{Replica\cite{straub2019replica}}} & \multicolumn{2}{c|}{\textbf{LERF-Mask\cite{GS-Grouping}}}  \\
\rowcolor{mygray}
\multirow{-2}{*}{\textbf{Method}} & \multirow{-2}{*}{\textbf{Venue}}  & \multirow{-2}{*}{\textbf{Type}} & \textbf{mIoU(\%)} & \textbf{F-score(\%)} & \textbf{mIoU(\%)} & \textbf{mBIoU(\%)}  \\ \hline
\hline
    \rowcolor{cyan!5}
GaussianGrouping\cite{GS-Grouping} & \pub{ECCV'24} & Pre-Process.   & 23.6 & 30.4 & 72.8 & 67.6  \\
OmniSeg3D\cite{OmniSeg3D} & \pub{CVPR'24} & Post-Process.
  & 39.1 & 35.9  & 74.7 & 71.8 \\ 
  \rowcolor{cyan!5}
Gaga\cite{Gaga} & \pub{ArXiv'24} & Pre-Process. &  - & - & 74.7 & 72.2 \\ 
Unified-Lift\cite{Unified-Lift} & \pub{CVPR'25} &End-to-End& \best{41.6} & \best{43.9} & \best{80.9} & \best{77.1} \\ 
\hline
\end{tabular}
}
\vspace{-5pt}
\label{tab:instance_seg}

\end{table}

\begin{table}[t]
\centering
\caption{\textbf{Quantitative 3D interactive segmentation on NVOS~\cite{ren2022neural} and SPIn-NeRF\cite{SPIn-NeRF}} in terms of mIoU and mAcc. } 

\vspace{-3mm}
\footnotesize
			    \setlength{\tabcolsep}{1mm}
			\renewcommand\arraystretch{1.0}
\resizebox{0.498\textwidth}{!}{
\begin{tabular}{|r||c|c|c|cc|cc|}
       \hline\thickhline
\rowcolor{mygray}
 & &  & &\multicolumn{2}{c|}{\textbf{NVOS\cite{ren2022neural}}} & \multicolumn{2}{c|}{\textbf{SPIn-NeRF\cite{SPIn-NeRF}}}   \\
\rowcolor{mygray}
\multirow{-2}{*}{\textbf{Method}} & \multirow{-2}{*}{\textbf{Venue}} & \multirow{-2}{*}{\textbf{Train}}& \multirow{-2}{*}{\textbf{Input Prompt}} & \textbf{mIoU(\%)} & \textbf{mAcc(\%)} & \textbf{mIoU(\%)} & \textbf{mAcc(\%)}  \\ \hline
\hline
\rowcolor{cyan!5}
\hline
\rowcolor{cyan!5}
SA3D-GS\cite{SA3D-GS} & \pub{NeurIPS'23} & \cmarkg&2D Click/Text & 90.7  & 98.3 & 93.2 & 99.1 \\ 

SAGD\cite{SAGD}  & \pub{ArXiv'24} & \xmark & 2D Click/Text & 90.4
 & 98.2 & 89.9  & 98.7 \\ 
\rowcolor{cyan!5}
FlashSplat\cite{FlashSplat} & \pub{ECCV'24} & \xmark & 2D Click
 & 91.8  & \best{98.6} & -  & -   \\ 

 Click-Gaussian\cite{Click-Gaussian} & \pub{ECCV'24} & \cmarkg&2D Click  & -  & - & \best{94.0} & -  \\ 
 \rowcolor{cyan!5}

 GaussianCut \cite{GaussianCut} & \pub{NeurIPS'24}& \xmark & 2D Click/Text
 & \best{92.5}  & {98.4} & 92.9  & \best{99.2}  \\ 
 SAGA\cite{SAGA} & \pub{AAAI'25} & \cmarkg & 2D Click/Text &  90.9  & 98.3 & 88.0 & 98.5 \\
\rowcolor{cyan!5}

 COB-GS\cite{COB-GS} & \pub{CVPR'25} &\cmarkg & Text
 & 92.1   & \best{98.6}   & - & - \\ 
iSegMan\cite{iSegMan} & \pub{CVPR'25} &\xmark & 2D Click & 92.0 &98.4 & 92.4 &99.1   \\
\rowcolor{cyan!5}

LUDVIG\cite{LUDVIG} & \pub{ICCV'25} &\xmark & Text & 92.4  &- & 93.8 &-   \\

\hline
\end{tabular}
}
\label{tab:interactive_seg}

\vspace{-5pt}
\end{table}

\begin{table}[!t]
\centering
\caption{\textbf{Performance comparison of open-vocabulary 3D semantic segmentation on ScanNet~\cite{dai2017scannet} benchmark.} The evaluation metrics are mIoU and mAcc, respectively.}
\vspace{-3mm}

\footnotesize
\resizebox{0.498\textwidth}{!}{
    \setlength{\tabcolsep}{1.2mm}
\begin{tabular}{|r||c|c|cc|cc|cc|}
       \hline\thickhline
\rowcolor{mygray}
& & &  \multicolumn{2}{c|}{\textbf{19 classes}} & \multicolumn{2}{c|}{\textbf{15 classes}} & \multicolumn{2}{c|}{\textbf{ 10 classes}}  \\
\rowcolor{mygray}
\multirow{-2}{*}{\textbf{Method}} & \multirow{-2}{*}{\textbf{Venue}} & \multirow{-2}{*}{\textbf{Evaluation}} & \textbf{mIoU} & \textbf{mAcc} & \textbf{mIoU} & \textbf{mAcc} & \textbf{mIoU} & \textbf{mAcc}  \\
\hline
\hline
\rowcolor{cyan!5}
LangSplat~\cite{LangSplat} & \pub{CVPR'24} & Point  &\ \ 3.8& \ \ 9.1 &\ \ 5.4 &13.2 &\ \ 8.4 &22.1\\
LEGaussians~\cite{LEGaussians}& \pub{CVPR'24} & Point   &\ \ 3.8 &10.9 &\ \ 9.0 &22.2& 12.8& 28.6\\
\rowcolor{cyan!5}
OpenGaussian\cite{OpenGaussian} & \pub{NeurIPS'24} & Point  & 24.7 &41.5& 30.1 &48.3& 38.3 &55.2\\
InstanceGaussian\cite{InstanceGaussian} & \pub{CVPR'25} & Point &40.7& 54.0 &\best{42.5}& \best{59.2}& \best{47.9}& \best{64.0}   \\
\rowcolor{cyan!5}
PanoGS\cite{PanoGS} & \pub{CVPR'25} & Point  & \best{50.7}& \best{70.2} & - & - & - & -   \\
COS3D\cite{COS3D} & \pub{NeurIPS'25}& Point  &32.4 &49.0 &35.9& 54.3 &44.3 &63.6\\
\hline
Dr. Splat\cite{DrSplat} & \pub{CVPR'25} & Gaussian  & 28.0 &44.6 &38.2 &60.4 &47.2 &68.9 \\
\rowcolor{cyan!5}
CAGS\cite{CAGS} & \pub{ArXiv'25} & Gaussian  &\best{32.6}& \best{48.9}& \best{41.1}& \best{62.0}& \best{54.8} &\best{75.9} \\

\hline
\end{tabular}
}
\vspace{-8pt}
\label{tab:scannet_ovs}
\end{table}

\noindent\textbf{SPIn-NeRF}~\cite{SPIn-NeRF} contains 10 real-world forward-facing scenes with annotated object masks. Each scene includes 60 training images with the object and 40 test images without the object.

\noindent\textbf{3D-OVS}~\cite{3D-OVS} is a benchmark for open-vocabulary 3D segmentation with language-aligned semantic labels, containing 10 scenes with around 30 360° images each.

\noindent\textbf{LERF-OVS}~\cite{LangSplat} contains open-vocabulary semantic annotations for 4 scenes derived from the LERF dataset~\cite{LERF}. 

\noindent\textbf{LERF-Mask}~\cite{GS-Grouping} contains semantic annotations of three scenes from LERF dataset~\cite{LERF} with a total of 23 prompts.

\noindent\textbf{Ref-LERF}\cite{Ref-LERF} extends the LERF dataset\cite{LERF} with 295 language expressions referring to 59 objects in 4 scenes, emphasizing spatial relations to support referring segmentation~\cite{ding2025multimodal,GRES} in 3DGS.

\noindent\textbf{SceneSplat-7K}\cite{SceneSplat} is the first large-scale, high-quality 3DGS dataset for indoor environments, containing 7,916 scenes aggregated from seven well-established datasets.

\noindent\textbf{SceneSplat-49K}\cite{ma2025scenesplat++} is a curated dataset of diverse indoor and outdoor scenes, featuring complex, high-quality full-scene 3DGS reconstructions from multiple sources.

\noindent$\bullet$~\textbf{Metrics.}
Segmentation performance in 3DGS is typically evaluated from three aspects: region accuracy, boundary quality, and instance-level quality. 

\noindent\textbf{Region accuracy.}
\textbf{Mean Intersection over Union (mIoU)} measures the average overlap ratio between predicted and ground-truth regions across classes, and is the most commonly used metric for semantic segmentation. \textbf{Mean Accuracy (mAcc)} measures the proportion of correctly classified pixels or instances, reflecting overall classification correctness, but it is generally less sensitive than mIoU to class imbalance or boundary errors. \textbf{F-score}, typically computed under an IoU threshold (e.g., 0.5), balances precision and recall and is often used as a complementary measure for detection- or instance-style evaluation.

\noindent\textbf{Boundary quality.}
\textbf{Mean Boundary Intersection over Union (mBIoU)} computes IoU over narrow boundary regions and is therefore more sensitive to contour quality than standard mIoU. This is particularly useful when evaluating fine-grained masks or object boundaries.

\noindent\textbf{Instance-level quality.}
\textbf{Panoptic Quality (PQ)}~\cite{kirillov2019panoptic} jointly measures recognition quality (RQ) and segmentation quality (SQ), and is more suitable when instance separation is important. 

\noindent
Overall, these metrics are complementary: mIoU and mAcc emphasize region correctness, mBIoU highlights boundary precision, while PQ and F-score are more informative for instance-aware settings.

\noindent$\bullet$~\textbf{Results.}
Following the taxonomy in Sec.~\textcolor{red}{2.2.1}, we report results across four representative segmentation settings.
For 3D instance segmentation (Table~\ref{tab:instance_seg}), Unified-Lift~\cite{Unified-Lift} achieves the best overall performance on both the Replica~\cite{straub2019replica} and LERF-Mask~\cite{GS-Grouping} datasets, demonstrating the effectiveness of consistency-aware end-to-end learning.
In the setting of 3D interactive segmentation (Table~\ref{tab:interactive_seg}), GaussianCut~\cite{GaussianCut} performs best on NVOS~\cite{NVOS}, while Click-Gaussian~\cite{Click-Gaussian} attains the highest mIoU on SPIn-NeRF~\cite{SPIn-NeRF}. Notably, GaussianCut is a training-free method that flexibly supports both 2D clicks and text prompts as input.
For open-vocabulary 3D semantic segmentation (Table~\ref{tab:scannet_ovs}), PanoGS~\cite{PanoGS} and InstanceGaussian~\cite{InstanceGaussian} achieve the best performance under point-based evaluation protocols, whereas CAGS~\cite{CAGS} performs best under 3D Gaussian-based evaluation.
In the open-vocabulary 2D semantic segmentation setting, VLGaussian~\cite{3DVision-LanguageGS} reaches a state-of-the-art mIoU of 97.1\% on the 3D-OVS benchmark~\cite{3D-OVS}, significantly outperforming existing approaches in Table~\ref{tab:3dovs}.
Table~\ref{tab:lerf_ovs} further presents results for two subtasks: open-vocabulary 2D localization (detection) and semantic segmentation. FMLGS~\cite{FMLGS} achieves the best performance on the segmentation task, while VLGaussian~\cite{3DVision-LanguageGS} leads in localization.
Across these settings, most methods leverage powerful vision-language foundation models such as CLIP and SAM to enhance open-vocabulary capabilities.

\begin{table}[t]
    \centering
    \caption{\textbf{Quantitative open-vocabulary 2D semantic segmentation on 3D-OVS~\cite{3D-OVS} dataset} in terms of mIoU.}
    \vspace{-3mm}
\footnotesize        \setlength{\tabcolsep}{1mm}
  \resizebox{0.498\textwidth}{!}{  \begin{tabular}{|r||c|c|cccccc|}
       \hline\thickhline
    
    \rowcolor{mygray}\textbf{Method} & \textbf{Venue} & \textbf{Foundation Model}  & \textbf{Bed} & \textbf{Bench} & \textbf{Room} & \textbf{Sofa} & \textbf{Lawn} & \textbf{Mean} \\
    \hline
    \hline
    \rowcolor{cyan!5}
    LangSplat\cite{LangSplat} & \pub{CVPR’24} & CLIP \& SAM &  92.5 & 94.2 & 94.1 & 90.0 & 96.1 & 93.4  \\
    Feature 3DGS\cite{Feature-3DGS} & \pub{CVPR’24}  & CLIP  & 83.5 & 90.7 & 84.7 & 86.9 & 93.4 & 87.8\\ 
    \rowcolor{cyan!5} 
    LEGaussians\cite{LEGaussians} & \pub{CVPR’24}  & CLIP  & 84.9 & 91.1 & 86.0 & 87.8 & 92.5 & 88.5\\
    GaussianGrouping\cite{GS-Grouping} & \pub{ECCV’24}  & SAM  & 83.0 & 91.5 & 85.9 & 87.3 & 90.6 & 87.7 \\
    \rowcolor{cyan!5}

    N2F2\cite{N2F2} & \pub{ECCV’24}  & CLIP \& SAM  & 93.8 & 92.6 & 93.5 & 92.1 & 96.3 & 93.9  \\
    GOI\cite{GOI} & \pub{MM’24}  & APE\cite{APE} \& CLIP  & 89.4 & 92.8 & 91.3 & 85.6 & 94.1 & 90.6 \\
    \rowcolor{cyan!5} 

    FMGS\cite{FMGS} & \pub{IJCV’24}  & CLIP \& DINO & 80.6 & 84.5 & 87.9 & 90.8 & 92.6 & 87.3  \\
    SLGaussian\cite{SLGaussian} & \pub{ArXiv'24} & CLIP \& SAM  & 41.3 & 47.0 & - & 30.3 & 54.4 & 44.1  \\
    \rowcolor{cyan!5} 
    FMLGS\cite{FMLGS} & \pub{ArXiv'25} & SAM2 \& CLIP & 95.7 & 96.3 & 96.8 & 95.2 & - & 96.0 \\
        LangSplatV2\cite{LangSplatV2}& \pub{NeurIPS'25} & CLIP \& SAM&93.0 &94.9&96.1 &92.3 &96.6 &94.6\\

    SAGA\cite{SAGA} & \pub{AAAI’25} & CLIP \& SAM  & {97.4} & 95.4 & 96.8 & 93.5 & 96.6 & 96.0  \\
      \rowcolor{cyan!5} 
  
    FastLGS\cite{FastLGS} & \pub{AAAI’25}  & CLIP \& SAM & 94.7 & 95.1 & 95.3 & 90.6 & 96.2 & 94.4 \\
    econSG\cite{econSG} & \pub{ICLR’25}  & SAM & 94.9 & 93.0 & 95.8 & 91.6 & 96.3 & 94.3  \\
    \rowcolor{cyan!5} 
    VLGaussian\cite{3DVision-LanguageGS} & \pub{ICLR'25}  & CLIP  & 96.8 & \best{97.3} & \best{97.7} & {95.5} & \best{97.9} & \best{97.1} \\
    LBG\cite{LBG} & \pub{WACV'25}  & CLIP \& SAM \& DINO  & {97.7} & {96.3} & {95.9} & \best{97.3} & {87.4} & {94.9}  \\
        \rowcolor{cyan!5} 

    CLIP-GS\cite{CLIP-GS} & \pub{TOMM'25} & CLIP \& SAM & 97.2& 94.8 & - & 94.1 & 96.5 &95.6 \\
    CCL-LGS\cite{CCL-LGS} & \pub{ICCV'25} & SAM \& CLIP & 97.3& 95.0 & -&92.3& 96.1& 95.2 \\
        \rowcolor{cyan!5} 

    ObjectGS~\cite{ObjectGS}& \pub{ICCV'25} & SAM2 &\best{98.0}& 96.4 &95.1 &97.2& 95.4&96.4\\
    \hline
    \end{tabular}
    }
\vspace{-5pt}
    \label{tab:3dovs}
\end{table}

\vspace{-5pt}

\begin{table}[t]
\centering
\caption{\textbf{Quantitative results on LERF-OVS~\cite{LangSplat} Dataset}. We report the open-vocabulary localization accuracy (mAcc) and 2D semantic segmentation (mIoU).}
\vspace{-3mm}

\footnotesize
\resizebox{0.498\textwidth}{!}{
    \setlength{\tabcolsep}{1mm}
\begin{tabular}{|r||c|cc|cc|cc|cc|}
       \hline\thickhline
       \rowcolor{mygray}
& &  \multicolumn{2}{c|}{\textbf{Ramen}} & \multicolumn{2}{c|}{\textbf{Figurines}} & \multicolumn{2}{c|}{\textbf{Teamtime}} & \multicolumn{2}{c|}{\textbf{Kitchen}} \\
\rowcolor{mygray}
\multirow{-2}{*}{\textbf{Method}} & \multirow{-2}{*}{\textbf{Venue}}  & \textbf{mIoU} & \textbf{mAcc} & \textbf{mIoU} & \textbf{mAcc} & \textbf{mIoU} & \textbf{mAcc} & \textbf{mIoU} & \textbf{mAcc} \\
\hline
\hline
\rowcolor{cyan!5}
LangSplat\cite{LangSplat} & \pub{CVPR'24}   & 51.2 & 73.2 & 44.7 & 80.4 & 65.1 & 88.1 & 44.5 & 95.5\\
Feature3DGS\cite{Feature-3DGS} & \pub{CVPR'24}   & 43.7 & 69.8 & 40.5 & 73.4 & 58.8 & 77.2 & 39.6 & 87.6  \\
\rowcolor{cyan!5}
LEGaussians\cite{LEGaussians} & \pub{CVPR'24}   & 46.0 & 67.5 & 40.8 & 75.2 & 60.3 & 75.6 & 39.4 & 90.3  \\
GaussianGrouping\cite{GS-Grouping} & \pub{ECCV'24}  & 45.5 & 68.6 & 40.0 & 74.3 & 60.9 & 75.0 & 38.7 & 88.2 \\
\rowcolor{cyan!5}
N2F2\cite{N2F2} & \pub{ECCV'24}   & 56.6 & 78.8 & 47.0 & 85.7 & 69.2 & 91.5 & 47.9 & 95.5 \\
FMGS\cite{FMGS} & \pub{IJCV'24}   & - & {90.0} & - & 93.8 & - & 89.7 & - & 92.6 \\
\rowcolor{cyan!5}
GOI\cite{GOI} & \pub{MM'24}   & 52.6& 75.5 & 44.5 & 82.9 & 63.7 & 88.6 & 41.4 & 90.4 \\
LangSurf\cite{LangSurf} & \pub{ArXiv'24}   & 47.0 & 63.4 & - & - & 73.6 & 84.8 & 55.0 & 81.9 \\
\rowcolor{cyan!5}
FMLGS\cite{FMLGS} & \pub{ArXiv'25}   & \best{73.2} & {89.2} & \best{72.4} & {94.3} & \best{81.8} & \best{96.7} & {64.3} & {96.2}  \\
FastLGS\cite{FastLGS} & \pub{AAAI'25}  &  - & 84.2 & - & 91.4 & - & 95.0 & - & 96.2   \\
\rowcolor{cyan!5}
econSG\cite{econSG} & \pub{ICLR'25}  & - & 83.2 & - & 89.3 & - & 93.4 & - & 96.2 \\
VLGaussian\cite{3DVision-LanguageGS} & \pub{ICLR'25}   & 61.4 & \best{92.5} & 58.1 & \best{97.1} & 73.5 & 95.8 & 54.8 & \best{98.6}  \\
\rowcolor{cyan!5}
LangScene-X\cite{LangScene-X} & \pub{ICCV'25}   & 42.9 & 72.7 & - & - & 45.0 & 78.9 & 63.6 & 90.9 
 \\
LUDVIG\cite{LUDVIG} & \pub{ICCV'25}  & 58.1 &  78.9& 63.3 & 80.4 & 77.1 & 94.9 & 58.5 & 90.9 \\
\rowcolor{cyan!5}
CCL-LGS\cite{CCL-LGS} & \pub{ICCV'25}  & 62.3& -& 61.2&-& 71.8&- & \best{67.1}&- \\

LangSplatV2\cite{LangSplatV2} & \pub{NeurIPS'25}&51.8 &74.7 &  56.4  &82.1 & 72.2& 93.2&59.1& 86.4 \\
ReLaGS\cite{ReLaGS} & \pub{CVPR'26}&51.2 &- &  64.7  &- & 81.0& -&60.6& - \\
\bottomrule
\end{tabular}}
\vspace{-8pt}
\label{tab:lerf_ovs}
\end{table}

\begin{table*}[t]
\centering
\caption{\textbf{Statistics of representative 3DGS editing datasets.} See Sec.~\textcolor{red}{4.2} for more detailed descriptions.}

\vspace{-3mm}
\begin{threeparttable}
\fontsize{10pt}{12pt}\selectfont
\resizebox{0.99\textwidth}{!}{
\setlength\tabcolsep{3pt}
\renewcommand\arraystretch{1.0}
\begin{tabular}{|r||c|c|c|l|}
       \hline\thickhline
\rowcolor{mygray}
\textbf{Dataset} & \textbf{Venue} & \textbf{\#Scenes} & \textbf{\#Views(Avg.)} & \textbf{Characterization} \\
\hline
\hline
\rowcolor{mygray2}

DTU~\cite{jensen2014large} &  \mypub{CVPR'14} & 80 &343 & Each scene consists of 49 or 64 accurate camera positions and reference structured light scans.\\
Tanks and Temples~\cite{knapitsch2017tanks} & \mypub{TOG'17} & 14 & - & It includes individual objects (\eg, ``Tank'', ``Train'') and large indoor scenes (\eg, ``Auditorium'', ``Museum''). \\
\rowcolor{mygray2}

GL3D~\cite{shen2018matchable} & \mypub{ACCV'18} & 543 & 230 & \makecell[l]{It contains 125,623 high-resolution images, most of which are captured by drones from multiple scales\\ and perspectives with significant geometric overlap. The images cover urban areas, rural regions, and scenic spots.} \\
LLFF~\cite{LLFF} & \mypub{TOG'19} & 32 & 25 & Using the COLMAP structure from motion implementation to compute poses for real images. \\
\rowcolor{mygray2}
NeRF-synthetic~\cite{NeRF} & \mypub{ECCV'20} & 8 & 100 & Each object is placed on a white background, with images at 800×800 resolution and corresponding camera poses. \\
BlendedMVS~\cite{yao2020blendedmvs} & \mypub{CVPR'20} &113& 158 & A large-scale MVS dataset, which contains a total of 17,818 images. \\
\rowcolor{mygray2}
Co3D~\cite{reizenstein2021common} & \mypub{CVPR'21} & - &- & \makecell[l]{Consists of 1.5 million frames extracted from approximately 19,000 videos, capturing objects from 50 MS-COCO\\ categories. Each image is annotated with camera poses and ground-truth 3D point clouds.}\\
Mip-NeRF360~\cite{barron2022mipnerf360unboundedantialiased} & \mypub{CVPR'22} & 9 & 215 & It consists of 360-degree panoramic images from both indoor and outdoor scenes. \\

\rowcolor{mygray2}
Nerfstudio~\cite{tancik2023nerfstudio} & \mypub{SIGGRAPH'23} & 10 & - & 4 phone captures with pinhole lenses and 6 mirrorless camera captures with a fisheye lens. \\
SPIn-NeRF~\cite{SPIn-NeRF} & \mypub{CVPR'23} & 10 & 100 & Providing challenging real-world scenes with views both with and without a target object.  \\
\rowcolor{mygray2}

IN2N~\cite{haque2023instructnerf2nerfediting3dscenes} & \mypub{ICCV'23} & 6 & 172 & Enabling structured and globally consistent 3D scene modifications while preserving the original scene's identity.  \\
ScanNet++~\cite{yeshwanthliu2023scannetpp} & \mypub{ICCV'23} &460 & 608& 280,000 captured DSLR images, and over 3.7M iPhone RGBD frames. \\
\rowcolor{mygray2}

360-USID~\cite{360-InpaintR} & \mypub{CVPR'25} & 7 &300 & Four outdoor (Box, Cone, Lawn, Plant) and three indoor (Cookie, Sunflower, Dustpan).\\

\hline
\end{tabular}
}
\end{threeparttable}
\label{tab:3dgs_editing_datasets}
\end{table*}

\begin{table*}[t]
    \centering
    \caption{\textbf{Performance comparison of 3DGS editing on Mip-NeRF360~\cite{barron2022mipnerf360unboundedantialiased} and IN2N~\cite{haque2023instructnerf2nerfediting3dscenes} datasets.} The similarity for CLIP Text-Image Direction,CLIP Text-Image, CLIP Image-Image, DINO are denoted as $\text{CLIP}_{dir}$, $\text{CLIP}_{T2I}$, $\text{CLIP}_{I2I}$, and $\text{DINO}_{sim}$, respectively.}
    \vspace{-3mm}
\footnotesize
    \resizebox{0.998\textwidth}{!}{
        \setlength{\tabcolsep}{1.5mm} 
    \begin{tabular}{|r||c|c|c|c|rrrrrrrr|}
       \hline\thickhline
    \rowcolor{mygray}
    \textbf{Methods} &\textbf{Venue}& \textbf{Category} & \textbf{Condition} &\textbf{Foundation Model} & \textbf{$\text{CLIP}_{dir}$ $\uparrow$ }& \textbf{$\text{CLIP}_{T2I}$ $\uparrow$} & \textbf{$\text{CLIP}_{I2I}$ $\uparrow$}& \textbf{$\text{DINO}_{sim}$ $\uparrow$}& \textbf{FID $\downarrow$} & \textbf{PSNR $\uparrow$}& \textbf{SSIM $\uparrow$}& \textbf{FPS $\downarrow$} \\

    \hline
    \hline
    \rowcolor{pink!30}
    IN2N~\cite{haque2023instructnerf2nerfediting3dscenes} &\pub{ICCV'23}&     \multicolumn{11}{c|}{\textbf{Baseline}}
\\
    \hline
    \rowcolor{cyan!5}
    GaussianEditor~\cite{GaussianEditor-HGS} &\pub{CVPR'24}& Localizing the Editing Object & Text&InstructPix2Pix~\cite{brooks2023instructpix2pix} & +29.44\% &- &- &- &- &- &- & -  \\
    GaussianEditor~\cite{GaussianEditor-ROI} &\pub{CVPR'24}& Localizing the Editing Object  & Text&InstructPix2Pix~\cite{brooks2023instructpix2pix} & +27.27\% &- &+11.76\% &- &\best{-50.49\%} &- &- & -60.87\% \\
    \rowcolor{cyan!5}
    TIP-Editor~\cite{TIP-Editor} & \pub{TOG'24}& Parameter-Efficient Tuning& Text\&Image&Stable Diffusion~\cite{rombach2022high}  & \best{+86.75\%} &- &- &+8.52\% &- &- &- & - \\
    GaussCtrl~\cite{GaussCtrl} & \pub{ECCV'24}& Multi-View Consistency   & Text&ControlNet~\cite{li2024controllabletextto3dgenerationsurfacealigned} & +19.34\% &- &- &- &- &- &- & -33.33\% \\
    \rowcolor{cyan!5}
    DGE~\cite{DGE} & \pub{ECCV'24}& Multi-View Consistency & Text &InstructPix2Pix~\cite{brooks2023instructpix2pix} & +4.69\% &+5.12\% &- &- &- &- &- & \best{-92.16\%} \\
    GaussianVTON~\cite{GaussianVTON}&\pub{ArXiv'24} & Multi-Stage Refinement   &Image&Stable Diffusion~\cite{rombach2022high} & - &\best{+105.81\%} &\best{+23.29\%} &- &-40.47\% &\best{+28.39\%} &\best{+11.61\%} & -  \\
    \rowcolor{cyan!5}
        Gomel~\etal~\cite{gomel2024diffusion} &\pub{ArXiv'24} & Multi-View Consistency& Text  &Stable Diffusion~\cite{rombach2022high} & +26.00\% &+9.09\% &+10.50\% &\best{+43.44\%} &- &+2.37\% &- & -  \\
    ProGDF~\cite{ProGDF}&\pub{ArXiv'24} & Efficiency and Speed & Text&InstructPix2Pix~\cite{brooks2023instructpix2pix}& +36.25\% &- &- &- &- &- &- & -  \\
    \rowcolor{cyan!5}
    TrAME~\cite{TrAME} &\pub{TMM'25}&  Multi-View Consistency  & Text & Stable Diffusion~\cite{rombach2022high} & - &- &+9.15\% &- &- &+16.60\% &- & -  \\
    DreamCatalyst~\cite{kim2024dreamcatalyst}&\pub{ICLR'25} & Efficiency and Speed&Text& InstructPix2Pix~\cite{brooks2023instructpix2pix}  & +10.14\% &- &+1.04\% &- &- &- &- & -46.15\%  \\

    \hline
    \end{tabular}
    }
    \vspace{-8pt}
    \label{tab:edit_comparision}
\end{table*}

\begin{table}[t]
    \centering
    \caption{\textbf{Performance comparison of 3DGS style transfer on Mip-NeRF360~\cite{barron2022mipnerf360unboundedantialiased} dataset,} evaluated by LPIPS and RMSE.}
    \vspace{-3mm}
\footnotesize
    \resizebox{0.492\textwidth}{!}{
    \setlength{\tabcolsep}{1mm}
    \begin{tabular}{|r||c|c|cccc|}
       \hline\thickhline
\rowcolor{mygray}
&  & & \multicolumn{2}{c}{\textbf{Short-Term Consis.}} & \multicolumn{2}{c|}{\textbf{Long-Term Consis.}}  \\
\rowcolor{mygray}
\multirow{-2}{*}{\textbf{Method}} & \multirow{-2}{*}{\textbf{Venue}}  & \multirow{-2}{*}{\textbf{2D Model}} & 
\textbf{{LPIPS}$\downarrow$} & \textbf{{RMSE}$\downarrow$} & \textbf{{LPIPS}$\downarrow$} & \textbf{{RMSE}$\downarrow$} \\
\hline
\hline
    \rowcolor{cyan!5}
    StyleGaus.~\cite{StyleGaussian} &\pub{SIGGRAPH'24} & VGG & 0.033 & 0.029 & 0.055 & 0.063  \\
    SemanticSplatSty.~\cite{SemanticSplatStylization} & \pub{TEGA'24}  & VGG & \best{0.019} & 0.042 & \best{0.028} & \best{0.055}   \\
    \rowcolor{cyan!5}
    InstantStyleGaus.~\cite{InstantStyleGaussian} &\pub{ArXiv'24} & Diffusion & 0.024 & \best{0.026} & 0.074 & 0.076  \\
    SGSST~\cite{galerne2024sgsst} & \pub{CVPR'25} & VGG &  0.030 &  0.032 & 0.055 & 0.063 \\
    \hline
    \end{tabular}
    }
    \vspace{-8pt}
    \label{tab:style_transfer_comparison}
\end{table}

\vspace{-1.16mm}
\subsection{ Performance Benchmarking: 3DGS Editing} \label{sec:exp_editing}

\noindent$\bullet$~\textbf{Datasets.}
We summarize the key features of the datasets in Table~\ref{tab:3dgs_editing_datasets}, with detailed descriptions provided below.

    \noindent\textbf{DTU}~\cite{jensen2014large} contains 80 scenes of large variability. Each scene consists of 49 or 64 accurate camera positions and reference structured light scans, all acquired by a 6-axis industrial robot.

    \noindent\textbf{Tanks and Temples}~\cite{knapitsch2017tanks} consists of 21 scenes, including individual objects such as ``Tank'' and ``Train,'' as well as large indoor scenes like ``Auditorium'' and ``Museum''.

    \noindent\textbf{GL3D}~\cite{shen2018matchable} contains 125,623 high-resolution images, mostly drone-captured at varying scales and angles with strong geometric overlap, covering urban, rural, and scenic areas.

    \noindent\textbf{LLFF}~\cite{LLFF} consists of both synthetic and real-world datasets. The synthetic part includes rendered images from 45,000 simplified indoor scenes, while the real-world part covers 24 training scenes and 8 testing scenes, each with 20–30 multi-view images.

    \noindent\textbf{NeRF-synthetic}~\cite{NeRF} consists of 8 scenes of an object placed on a white background. Each scene includes 100 training images with a resolution of 800×800 and associated camera poses.

    \noindent\textbf{BlendedMVS}~\cite{yao2020blendedmvs} contains 113 scenes with 20–1000 images each, totaling 17,818 images.

    \noindent\textbf{Co3D}~\cite{reizenstein2021common} contains 1.5 million frames from ~19,000 videos of objects across 50 MS-COCO categories, with camera poses and 3D point cloud annotations.

    \noindent\textbf{Mip-NeRF360}~\cite{barron2022mipnerf360unboundedantialiased} provides 360° indoor and outdoor panoramas for evaluating neural rendering methods, focusing on complex lighting, geometry, and textures.
    
    \noindent\textbf{Nerfstudio}~\cite{tancik2023nerfstudio} contains 10 scenes: 4 captured by phones with pinhole lenses and 6 by mirrorless cameras with fisheye lenses.

    \noindent\textbf{SPIn-NeRF}~\cite{SPIn-NeRF} contains 10 forward-facing in-the-wild scenes, including 3 indoor and 7 outdoor scenes, each with 100 multi-view images and annotated foreground masks.

    \noindent\textbf{IN2N}~\cite{haque2023instructnerf2nerfediting3dscenes} contains 6 real-world scenes (\eg, bear, face, person), each with an average of 172 images.

    \noindent\textbf{ScanNet++}~\cite{yeshwanthliu2023scannetpp} contains 460 scenes, 280,000 captured DSLR images, and over 3.7M iPhone RGBD frames.

    \noindent\textbf{360-USID}~\cite{360-InpaintR} includes 4 outdoor and 3 indoor scenes, each with 171–347 training and 31–33 novel views.

\noindent$\bullet$~\textbf{Metrics.}
Evaluation in 3DGS editing is inherently multi-dimensional, since a successful edit should ideally preserve scene identity and geometry while also satisfying the editing instruction. Accordingly, existing metrics can be grouped into four categories: text-alignment metrics, content-preservation metrics, image-fidelity metrics, and aesthetics and human evaluation.

\noindent\textbf{Text alignment.}
\textbf{CLIP Text-Image Similarity}~\cite{CLIP} measures how well the edited image matches the target text prompt. \textbf{CLIP Text-Image Direction Similarity}~\cite{CLIP} further evaluates whether the change from the original image to the edited image follows the semantic direction specified by the text instruction, making it particularly useful for instruction-guided editing.

\noindent\textbf{Content preservation.}
\textbf{CLIP Image-Image Similarity}~\cite{CLIP} and \textbf{DINO Similarity}~\cite{zhang2022dino} assess how well the edited result preserves the original subject or scene content. These metrics are helpful for measuring identity preservation, although they may not fully reflect whether the edit is semantically successful.

\noindent\textbf{Image-level fidelity.}
\textbf{Peak Signal-to-Noise Ratio (PSNR)}~\cite{PSNR}, \textbf{Structural Similarity (SSIM)}~\cite{SSIM}, and \textbf{RMSE} mainly measure pixel-level fidelity with respect to reference images, while \textbf{Learned Perceptual Image Patch Similarity (LPIPS)}~\cite{zhang2018unreasonable} is more correlated with perceptual similarity in deep feature space. \textbf{Frechet Inception Distance (FID)}~\cite{heusel2017gans} compares the distribution of generated images against real images and is more suitable for measuring overall realism at the dataset level rather than edit faithfulness for individual samples.

\noindent\textbf{Aesthetics and human evaluation.}
\textbf{Aesthetic score}~\cite{aesthetic-predictor} reflects subjective visual appeal, while \textbf{user study} is often necessary to evaluate aspects such as instruction faithfulness, scene consistency, and user preference that may not be well captured by automatic metrics.

\noindent
Overall, no single metric is sufficient for 3DGS editing. CLIP-based scores emphasize semantic alignment, image-similarity metrics favor preservation, and pixel-/perceptual metrics reflect fidelity, so most works require multiple complementary metrics for a meaningful evaluation.

\noindent$\bullet$~\textbf{Results.} For 3D image editing tasks, the most commonly used datasets are Mip-NeRF360~\cite{barron2022mipnerf360unboundedantialiased} and IN2N~\cite{haque2023instructnerf2nerfediting3dscenes}. We adopt these two datasets as benchmarks to record the performance of various methods. 
However, as most existing 3D editing methods are evaluated on different scene subsets, it remains difficult to perform a comprehensive and fair comparison under a unified protocol. Given that IN2N~\cite{haque2023instructnerf2nerfediting3dscenes} is widely adopted as a baseline, we report the relative gains of each method over IN2N to enable an indirect yet relatively fair comparison, as summarized in Table~\ref{tab:edit_comparision}.
Overall, GaussianVTON\cite{GaussianVTON} demonstrates the best performance across multiple metrics.
For the style transfer task, we use Mip-NeRF 360~\cite{barron2022mipnerf360unboundedantialiased} as the benchmark and compare the results of different methods in Table~\ref{tab:style_transfer_comparison}, where SemanticSplatSty.\cite{SemanticSplatStylization} achieves the best overall performance.

\begin{table*}[t]
\centering
\caption{\textbf{Statistics of representative 3DGS generation datasets.} See Sec.~\textcolor{red}{4.3} for more detailed descriptions.}\vspace{-3mm}

\begin{threeparttable}
\fontsize{10pt}{12pt}\selectfont
\resizebox{0.99\textwidth}{!}{
\setlength\tabcolsep{8pt}
\renewcommand\arraystretch{1.0}
\begin{tabular}{|r||c|c|c|l|}
       \hline\thickhline
\rowcolor{mygray}
\textbf{Dataset} & \textbf{Venue} & \textbf{\#Type} & \textbf{\#Scenes} & \textbf{Characterization} \\
\hline
\hline
\rowcolor{mygray2}
NYUdepth~\cite{silberman2012indoor} & \mypub{ECCV'12} & Image-to-3D & 464 & It contains 1449 RGB-D images, capturing 464 diverse indoor scenes with detailed annotations. \\
ShapeNet~\cite{chang2015shapenet} & \mypub{ArXiv'15} & Image\&Text-to-3D & 60k & These 3D models span 55 categories, each with a geometry file and unique identifier.  \\
\rowcolor{mygray2}
ScanNet~\cite{dai2017scannet} & \mypub{CVPR'17} & Image-to-3D & 1513 & It contains 2.5 M views in 1513 indoor scenes annotated with 3D camera poses.\\
RealEstate10K~\cite{zhou2018stereo} & \mypub{SIGGRAPH'18} & Image-to-3D & 80k & It comprises home walkthrough videos from YouTube.  \\
\rowcolor{mygray2}
Replica~\cite{straub2019replica}& \mypub{ArXiv'19} & Image-to-3D & 18 & A 3D indoor scene dataset featuring dense meshes, HDR textures, and semantic labels. \\
ACID~\cite{liu2021infinite} & \mypub{ICCV'21} & Image-to-3D & 13047 & Featuring aerial landscape videos, includes 11,075 training scenes and 1,972 testing scenes.\\
\rowcolor{mygray2}
GSO~\cite{GSO} & \mypub{ICRA'22} & Image\&Text-to-3D & 1030 & It comprises 3D scanned common household items.  \\
LAION-5B~\cite{schuhmann2022laion} & \mypub{NeurIPS'22} & Text-to-3D & - & LAION-5B's key feature is its vast scale, with 5.85 billion image-text pairs. \\
\rowcolor{mygray2}
Objaverse~\cite{deitke2023objaverse} & \mypub{CVPR'23} & Image\&Text-to-3D & 800K & Objaverse has vast scale of 800K+ 3D models with rich annotations.  \\
OmniObject3D~\cite{wu2023omniobject3d} & \mypub{CVPR'23} & Image-to-3D & 6k & A large-scale collection of high-quality real-scanned 3D objects with rich 2D and 3D annotations. \\
\rowcolor{mygray2}
LOM~\cite{cui2024aleth} & \mypub{AAAI'24} & Image-to-3D & 5 & It includes five real-world scenes, each with 25$\sim$48 sRGB images captured in difficult lighting. \\
G-objaverse~\cite{zuo2024high} & \mypub{ECCV'24} & Image\&Text-to-3D &280K& 10 general classes which gives about 280K samples.  \\
\rowcolor{mygray2}
DL3DV-10K~\cite{DL3DV-10K} &\mypub{CVPR'24}& Image-to-3D &10K& Large-scale scene dataset that contains both indoor and outdoor scenarios.\\

\hline
\end{tabular}
}
\end{threeparttable}
\label{tab:3dgs_generation_datasets}
\end{table*}
\begin{table*}[t!]
    \centering
    \caption{\textbf{Quantitative 3DGS generation on GSO~\cite{GSO} dataset.} CLIP Image-Image Similarity as {$\text{CLIP}_{I2I}$}.}
    \vspace{-3mm}
\footnotesize
    \resizebox{0.996\textwidth}{!}{
    \setlength{\tabcolsep}{3mm}
    
    \begin{tabular}{|r||c|c|c|c|cccccc|}
       \hline\thickhline
    
\rowcolor{mygray}
\textbf{Methods} & \textbf{Venue} & \textbf{Category} & \textbf{Condition} & \textbf{2D Foundation Model} &\textbf{PSNR$\uparrow$}&\textbf{SSIM$\uparrow$}&\textbf{LPIPS$\downarrow$}&\textbf{$\text{CLIP}_{I2I}$ $\uparrow$}&\textbf{FID$\downarrow$}&\textbf{KID$\downarrow$}\\
    \hline
        \hline
    \rowcolor{pink!30}
    \multicolumn{11}{c}{\textbf{Per-Scene Optimization}} \\
    \hline
    \rowcolor{cyan!5}
    DreamGaussian~\cite{Dreamgaussian} &\pub{ICLR'24}& Standard SDS  & Text\&Image &Stable Diffusion~\cite{rombach2022high}& 18.27 & 0.834 & 0.189 & 0.748 & -& - \\
    Hritam~\etal~\cite{basak2024enhancingsingleimage3d} &\pub{ArXiv'24}& Standard SDS & Image &Stable Diffusion~\cite{rombach2022high}& 22.16 & 0.887 & 0.121 & - & -& - \\
    \rowcolor{cyan!5}
    GECO~\cite{GECO} & \pub{ArXiv'24}& Improving SDS & Image &Zero123++~\cite{shi2023zero123++}&19.31 & 0.825 & 0.154 & - & -& -  \\
    
    \rowcolor{pink!30}
        \hline
    \multicolumn{11}{c}{\textbf{Feed-Forward-based}} \\
    \hline
    \rowcolor{cyan!5}
    LGM~\cite{LGM} & \pub{ECCV'24} & Multi-View-based  & Text\&Image &MVDream~\cite{shi2023mvdream}& 17.13 & 0.810 & 0.250 & - & \best{19.93}& \best{0.55}\\
    GRM~\cite{GRM} &\pub{ECCV'24}& Network Design  & Text\&Image &Zero123++~\cite{shi2023zero123++}& 25.03 & 0.899 & 0.102 & 0.869 & - & - \\
    \rowcolor{cyan!5}
        GS-LRM~\cite{GS-LRM} &\pub{ECCV'24}& Multi-View-based  & Image & Zero123++~\cite{shi2023zero123++} & 17.70 & 0.795 & 0.241 & - & 112.96& - \\

    Lu~\etal~\cite{lu2024large}&\pub{MM'24} & Network Design  & Image &-& 17.92 & 0.810 & 0.210 & - & -& -  \\
        \rowcolor{cyan!5}

    Hi3D~\cite{Hi3D} &\pub{MM'24}& Optimization Diffusion  & Image &Stable Video Diffusion~\cite{blattmann2023stable}& 24.26 & 0.864 & 0.119 & - & -& - \\
    DiffusionGS~\cite{cai2024bakinggaussiansplattingdiffusion}&\pub{ArXiv'24} & Optimization Diffusion  & Image&FLUX~\cite{flux2024} \& SD~\cite{rombach2022high} & 22.07 & 0.854 & 0.111 & - & 11.52& - \\
    \rowcolor{cyan!5}

    GeoGS3D~\cite{GeoGS3D}&\pub{ArXiv'24} & Multi-View-based & Image &Zero-1-to-3~\cite{liu2023zero}& 22.98 & 0.899 & 0.146 & - & -& -   \\
    NovelGS~\cite{NovelGS} &\pub{ArXiv'24}& Optimization Diffusion  & Text\&Image &-&\best{31.30} & \best{0.946} & \best{0.065} & - & -& -  \\
    \rowcolor{cyan!5}
    Cycle3D~\cite{Cycle3D} &\pub{AAAI'25}& Optimization Diffusion  & Image& MVDream~\cite{shi2023mvdream} & 21.40 & 0.884 & 0.115 & 0.855 & -& -\\
    GaussianAnything~\cite{GaussianAnything} &\pub{ICLR'25}&  Latent Optimization  & Text\&Image &-& - & - & - & - & 24.21& 0.76  \\
    \rowcolor{cyan!5}
    Flex3D~\cite{han2024flex3d} &\pub{ICML'25}& Multi-View-based  & Image&Emu Model~\cite{dai2023emu} &25.55 & 0.894 & 0.074 & \best{0.893} & -& -  \\
    Ouroboros3D~\cite{wen2024ouroboros3d} &\pub{CVPR'25}& Optimization Diffusion  & Image &Stable Video Diffusion~\cite{blattmann2023stable}& 21.76 & 0.889 & 0.109 & - & -& -\\
    \hline
    \end{tabular}
    }
    \vspace{-5pt}
    \label{tab:perform_gso}
\end{table*}

\begin{table*}[t!]
    \centering
    \caption{\textbf{Quantitative 3DGS generation on Objaverse~\cite{deitke2023objaverse} dataset}. CLIP Text-Image Similarity as $\text{CLIP}_{T2I}$.}
    \vspace{-3mm}
\footnotesize
    \resizebox{0.996\textwidth}{!}{
    \setlength{\tabcolsep}{3mm}
    \begin{tabular}{|r||c|c|c|c|cccccc|}
       \hline\thickhline
    \rowcolor{mygray}\textbf{Methods} & \textbf{Venue} & \textbf{Category} & \textbf{Condition} & \textbf{2D Foundation Model} &\textbf{PSNR$\uparrow$}&\textbf{SSIM$\uparrow$}&\textbf{LPIPS$\downarrow$}&\textbf{$\text{CLIP}_{T2I}$ $\uparrow$}&\textbf{FID$\downarrow$}&\textbf{KID$\downarrow$}\\
    \hline
    \hline
    \rowcolor{pink!30}
    \multicolumn{11}{c}{\textbf{Feed-Forward-based}} \\
    \hline
    \rowcolor{cyan!5}
    LGM~\cite{LGM} &\pub{ECCV'24}& Multi-View-based  & Text\&Image&MVDream~\cite{shi2023mvdream}& - & - &-& 27.21 & 123.8& 4.53  \\
    GVGEN~\cite{GVGEN} &\pub{ECCV'24}& Optimization Diffusion  & Text & Stable Diffusion~\cite{rombach2022high}& - &- & - & 27.33 & 132.4& 6.04  \\
    \rowcolor{cyan!5}
    GeoGS3D~\cite{GeoGS3D} &\pub{ArXiv'24}& Multi-View-based  & Image &Zero-1-to-3~\cite{liu2023zero}& \best{23.97} & \best{0.921} & \best{0.113} & - & -& - \\
    Wizadwongsa~\etal~\cite{wizadwongsa2024taming} &\pub{ArXiv'24}&  Latent Optimization  & Text&Stable Diffusion 3~\cite{esser2024scaling} &- & - & - & 27.61 & -& - \\
    \rowcolor{cyan!5}
    Atlas-Gaussians~\cite{Atlas-Gaussians} &\pub{ICLR'25}& Latent Optimization  & Text &Latent Diffusion Model~\cite{rombach2022high}& - &- & - & \best{30.66} & \best{109.5}& \best{4.04}\\
    Turbo3D~\cite{hu2024turbo3d} &\pub{CVPR'25}&  Latent Optimization  & Text&DiT~\cite{peebles2023scalable} & - & - & - & 27.61 & -& -  \\

    \hline
    \end{tabular}
    }
    \vspace{-8pt}
    \label{tab:perform_objaverse}
\end{table*}

\vspace{-1.2mm}
\subsection{Performance Benchmarking: 3DGS Generation} \label{sec:exp_generation}
\noindent$\bullet$~\textbf{Datasets.} We summarize the key features of the datasets in Table~\ref{tab:3dgs_generation_datasets}, with detailed descriptions provided below.

    \noindent\textbf{NYUdepth}~\cite{silberman2012indoor} contains 1449 RGBD images, capturing 464 diverse indoor scenes, with detailed annotations.

    \noindent\textbf{ShapeNet}~\cite{chang2015shapenet} contains approximately 50,000 3D models across 55 object categories. Each model includes a corresponding geometry file and a unique identifier.

    \noindent\textbf{ScanNet}~\cite{dai2017scannet} is a large RGB-D dataset containing 2.5 M views in 1,513 indoor scenes annotated with 3D camera poses.

    \noindent\textbf{RealEstate10K}~\cite{zhou2018stereo} contains 67,477 training and 7,289 testing home walkthrough video scenes from YouTube.

    \noindent\textbf{Replica}~\cite{straub2019replica} is a dataset of 18 highly photorealistic 3D indoor scenes, each with dense meshes, HDR textures, semantic and instance labels, and reflective surfaces like mirrors and glass.

    \noindent\textbf{ACID}~\cite{liu2021infinite} featuring aerial landscape videos, includes 11,075 training scenes and 1,972 testing scenes.
    
    \noindent\textbf{GSO}~\cite{GSO} comprises 1,030 3D scanned household items.

    \noindent\textbf{LAION-5B}~\cite{schuhmann2022laion} consists of three subsets: 2.32 billion English image-text pairs, 2.26 billion image-text pairs in over 100 other languages, and 1.27 billion samples with undetectable language. 

    \noindent\textbf{Objaverse}~\cite{deitke2023objaverse} is a large-scale dataset of objects with 800K+ 3D models with descriptive captions, tags, and animations.

    \noindent\textbf{OmniObject3D}~\cite{wu2023omniobject3d} contains 6,000 scanned objects across 190 daily categories. Each object includes textured meshes, point clouds, multiview renders, and real-captured videos.

    \noindent\textbf{LOM}~\cite{cui2024aleth} comprises five real-world scenes (``buu'', ``chair'', ``sofa'', ``bike'', and ``shrub''), each containing 25$ \sim $48 sRGB images captured by a DJI Osmo Action 3 camera under adverse lighting conditions, including low light and overexposure.

    \noindent\textbf{G-objaverse}~\cite{zuo2024high}. Derived from Objaverse~\cite{deitke2023objaverse}, G-Objaverse filters out poorly captioned 3D models and includes high-quality renderings produced via a hybrid of rasterization and path tracing.

    \noindent\textbf{DL3DV-10K}~\cite{DL3DV-10K} comprises 51.2M frames from 10,510 videos across 65 locations, covering diverse indoor and outdoor scenes with varying reflection, transparency, and lighting conditions.

\noindent$\bullet$~\textbf{Metrics.}
Evaluation in 3DGS generation usually involves several complementary aspects, including perceptual image quality, distributional realism, geometry-level quality, and human-centered evaluation.

\noindent\textbf{Perceptual image quality.}
\textbf{Natural Image Quality Evaluator (NIQE)}~\cite{mittal2012making} and \textbf{Blind/Referenceless Image Spatial Quality Evaluator (BRISQUE)}~\cite{mittal2011blind} are no-reference image quality metrics that estimate the naturalness of rendered results based on image statistics. They are useful for coarse perceptual assessment, but they do not explicitly measure semantic correctness or 3D consistency.

\noindent\textbf{Distributional realism.}
\textbf{Kernel Inception Distance (KID)}~\cite{binkowski2018demystifying}, \textbf{Inception Score}~\cite{salimans2016improved}, and \textbf{CMMD}~\cite{jayasumana2024rethinking} compare generated images against reference distributions in feature space. These metrics are more suitable for evaluating the realism and diversity of a set of generated images, but they may not directly reflect view consistency or geometric faithfulness.

\noindent\textbf{Geometry-level quality.}
\textbf{Chamfer Distance}~\cite{barrow1977parametric} measures similarity between generated and reference point sets, and is therefore more informative for geometric accuracy than purely image-based metrics. \textbf{Thresholded Symmetric Epipolar Distance (TSED)}~\cite{TSED} evaluates multi-view or sequential consistency by measuring the number of geometrically consistent frame pairs.

\noindent\textbf{Semantic and human-centered evaluation.}
\textbf{MLLM-based evaluation} (\eg, GPT-4, LLaVA~\cite{liu2023visual}) and \textbf{HPSv2}~\cite{wu2023human} provide higher-level assessments of semantic quality or human preference, while \textbf{user study} remains important for judging appearance quality, geometric plausibility, and subjective preference in ways that automatic metrics may miss.

\noindent
Overall, image-quality metrics, geometry metrics, and human-centered metrics capture different dimensions of generation performance. Strong scores on one class of metrics do not necessarily imply strong performance on the others, so a balanced evaluation typically requires several complementary criteria.

\noindent$\bullet$~\textbf{Results.} 
We select two representative datasets, GSO~\cite{GSO} and Objaverse~\cite{deitke2023objaverse}, as benchmarks for 3DGS generation. The performance of various methods is summarized in Table~\ref{tab:perform_gso} and Table~\ref{tab:perform_objaverse}, respectively.
Due to the limited number of publicly comparable baselines, we provide a quantitative assessment based on the available results.
On GSO dataset, NovelGS~\cite{NovelGS} achieves the best performance by tailoring the diffusion process to suit the 3D Gaussian representation. It supports both text-to-3D and image-to-3D generation, demonstrating strong generalization across input modalities.
On Objaverse dataset, Atlas-Gaussians~\cite{Atlas-Gaussians} performs best, leveraging latent space optimization to enhance generation quality. It primarily supports text-conditioned generation and shows promising results on open-domain content.

\renewcommand{\thesection}{A.\arabic{section}}
\section{{Functional Application Tasks}}\label{sec:3.4}

\subsection{Human Avatar}\label{sec:3.4.1}
Human avatars, digital representations of users in virtual spaces, enable immersive interaction across gaming, virtual meetings, and the metaverse. 3DGS-based avatar modeling can be divided into two main directions:
1) Body-based avatars~\cite{Animatable-gaussians,Hifi4g,chen2024ggavatar,TaoAvatar,kocabas2024hugs,abdal2024gaussian} typically rely on parametric body models like SMPL~\cite{SMPL} or SMPL-X~\cite{SMPLX} to guide canonical Gaussian initialization. Methods such as HUGS~\cite{kocabas2024hugs} and Animatable Gaussians~\cite{Animatable-gaussians} extend beyond rigid skeletons by introducing pose-conditioned deformation to handle garments and fine motions. Others like D3GA~\cite{D3GA} and GauHuman~\cite{Gauhuman} enhance realism with cage-based deformation or efficient monocular reconstruction to move beyond rigid templates. Despite progress, body-based methods still face challenges in handling loose clothing, fast motions, and generalization to unseen scenes.
2) Head-based avatars~\cite{GaussianAvatars,GPAvatar,HRAvatar,StrandHead} emphasize fine-grained reconstruction of facial geometry, expressions, and speech-driven dynamics. Many approaches combine FLAME~\cite{FLAME} with deformable Gaussians to capture subtle expressions. For instance, Gaussian Head Avatar~\cite{Gaussianheadavatar} replaces traditional LBS with MLP-based displacement, while FlashAvatar~\cite{FlashSplat} achieves 300 FPS real-time rendering. Despite progress, Head-based methods still face challenges in handling subtle facial expressions, uneven illumination, and low-quality pose.

\subsection{SLAM}\label{sec:3.4.2}
Simultaneous Localization and Mapping (SLAM) is a foundational task in robotics and vision, aiming to estimate camera poses and reconstruct 3D environments simultaneously.
We broadly categorize 3DGS-based SLAM~\cite{DynaGSLAM,cheng2025outdoor,SEGS-SLAM,Wildgs-slam} into two directions: geometry-based SLAM and semantics-aware SLAM. The former includes RGB-D SLAM and RGB SLAM, depending on depth availability. RGB-D SLAM methods (\eg, GS-SLAM~\cite{yan2024gs}, SplaTAM~\cite{keetha2024splatam}) leverage accurate depth to construct dense Gaussian maps and enable reliable tracking via silhouette rendering or Gaussian matching. In contrast, RGB SLAM (\eg, MonoGS~\cite{matsuki2024gaussian}, Photo-SLAM~\cite{Photo-slam}) must infer geometry through multi-view optimization or depth prediction, posing challenges in outdoor or dynamic scenes due to less stable depth cues.
Beyond geometry, semantic SLAM incorporates scene understanding to enable richer 3D maps for downstream tasks like navigation and interaction. Early works like SGS-SLAM~\cite{Sgs-slam} embed semantic colors into Gaussians, while more recent methods (\eg, OpenGS-SLAM~\cite{yang2025opengs}, GS$^3$LAM~\cite{li2024gs3lam}, LEGS~\cite{legs}) introduce learned semantic features or language-guided Gaussians to enhance high-level reasoning. Such semantic integration improves robustness, particularly under sensor noise or ambiguous geometry.
Despite progress, open challenges remain, like SLAM under sparse views, scalable memory, and consistent semantics in large-scale scenes.

\ifCLASSOPTIONcaptionsoff
  \newpage
\fi

\vspace{-0.7mm}
{
\bibliographystyle{IEEEtran}
\bibliography{IEEEabrv,egbib}
}

\end{document}